%% file: main.tex
\documentclass[10pt,twocolumn,letterpaper]{article}

\usepackage[pagenumbers]{cvpr} 

\input{macros}

\usepackage{cuted}  
\definecolor{cvprblue}{rgb}{0.21,0.49,0.74}
\usepackage[pagebackref,breaklinks,colorlinks,allcolors=cvprblue]{hyperref}

\def\paperID{*****} 
\def\confName{CVPR}
\def\confYear{2027}

\title{\emoji{joker}CARD: Contact-Anchored Retargeting for Dexterous Manipulation}

\author{Cunxi Dai$^{1,2}$ \quad Botao He$^{3,2\ast}$ \quad Francois R Hogan$^{2\ast}$ \quad Wei Pan$^{2\ast}$ \quad Rao Fu$^{4,2}$ \quad Wen-Han Hsieh$^{1}$ \\[2pt]
Ruoshi Liu$^{3,2}$ \quad Jitendra Malik$^{4,2}$ \quad Guanya Shi$^{1,2\dagger}$ \quad Akash Sharma$^{2\dagger}$ \\[6pt]
{\normalsize $^{1}$Carnegie Mellon University \quad $^{2}$Amazon Frontier AI \& Robotics (FAR)} \\[1pt]
{\normalsize $^{3}$University of Maryland, College Park \quad $^{4}$UC Berkeley} \\[2pt]
{\normalsize $^{\ast}$Core contributor \quad $^{\dagger}$Equal advising} \\[2pt]
{\normalsize Project page: \url{https://card-retargeting.github.io/}}
}
\newcommand{\ours}{\texttt{CARD}\xspace}
\newcommand{\fca}{\emph{contact force targets}\xspace}
\newcommand{\fcas}{\emph{contact force target}\xspace}

\definecolor{figleft}{HTML}{E8731F}
\definecolor{figcenter}{HTML}{27996A}
\definecolor{figright}{HTML}{0E8FA3}
\definecolor{figbottom}{HTML}{E5342A}

\begin{document}
\maketitle

\begin{strip}
    \centering
    \vspace{-1.5em}
    \includegraphics[width=0.9\linewidth]{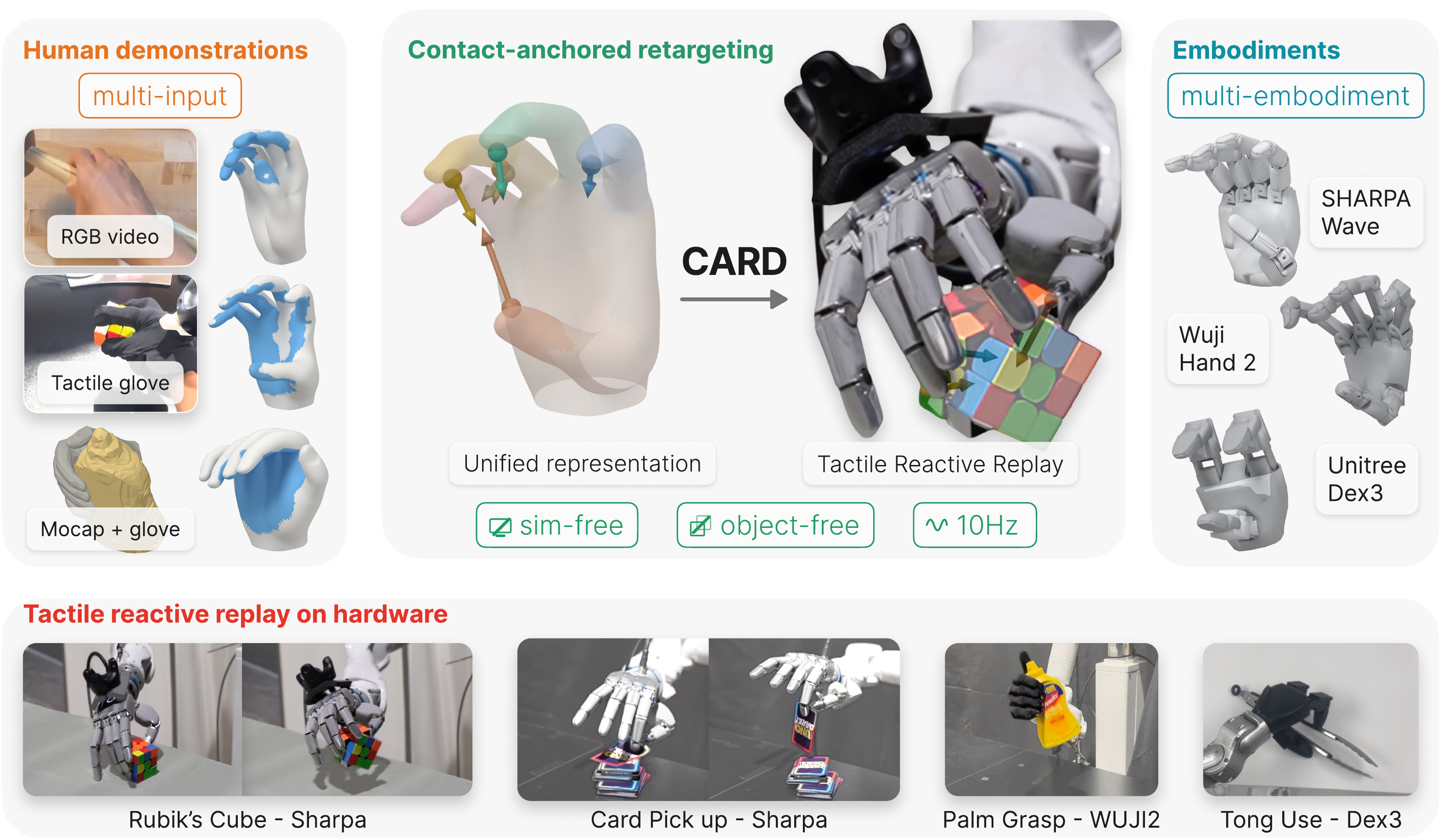}
    \vspace{4pt}
    \captionsetup{hypcap=false}
    \captionof{figure}{
    \textbf{\ours is a unified human-to-robot, multi-embodiment tactile retargeting framework.} Demonstrations from monocular video, tactile gloves, or motion capture with a tactile glove (\textcolor{figleft}{left}) map into a unified hand--contact representation that \ours retargets to robot joint commands and \fca  (\textcolor{figcenter}{center}). Our approach can retarget to different embodiments by changing only the semantic mapping (\textcolor{figright}{right}). Finally, a tactile-tracking controller replays open-loop on hardware (\textcolor{figbottom}{bottom}) while tracking the desired \fca, requiring no object model or simulator.}
    \captionsetup{hypcap=true}
    \label{fig:teaser}
\end{strip}


\begin{abstract}


Humans are dexterous, but traditional data collection paradigms such as teleoperation, kinesthetic teaching, and wearable exoskeletons force humans to act through the robot embodiment, bottlenecking their dexterity. Retargeting can avoid this. However, existing methods purely translate kinematics and ignore the \emph{real-world} contact dynamics essential for task success. We present \ours: Contact-Anchored Retargeting for Dexterous Manipulation, a simulator and object-free framework that retargets human demonstrations to robot actions by explicitly preserving \fca. \ours unifies human data sources including video, motion capture, and tactile gloves via a shared hand--contact representation and generates explicit \fca in addition to kinematics. To maintain these contact targets during hardware open-loop replay, \ours also introduces a tactile tracking controller. 
For video which does not contain an explicit tactile source, we present \emph{Vid-HACO}, our video-to-contact model that sets a new state of the art. Across our video, glove, motion-capture data and the GRAB dataset, \ours maintains contact on $43$--$78\%$ of contact frames, while kinematic and object-aware retargeting methods maintain contact on $<10\%$ and $22.5\%$ of the frames respectively. 
Further, integrating our formulation into physics-based retargeting improves strict task success from 12\% to 32\% on GRAB~\cite{grab}. Finally, we show that our tactile tracker is critical to reproduce demonstrated forces during hardware replay. These results establish that contact preservation via \ours is a scalable interface for transferring human dexterity to robots.

\end{abstract}


\input{sections/introduction}
\input{sections/related_work}
\input{sections/method}
\input{sections/experiments}
\input{sections/conclusion}



{
    \small
    \bibliographystyle{ieeenat_fullname}
    \bibliography{references}
}


\clearpage
\setcounter{page}{1}
\maketitlesupplementary
\appendix
\crefalias{section}{appendix}
\crefalias{subsection}{appendix}

\input{sections/appendix}

\end{document}

%% file: macros.tex
\usepackage[T1]{fontenc}
\usepackage[utf8]{inputenc}
\usepackage{mathtools}
\usepackage{microtype}
\usepackage{inconsolata}
\usepackage{xstring}
\usepackage[hang,flushmargin]{footmisc}
\usepackage{iftex}
\ifLuaTeX
  \usepackage{emoji}
\else
  \providecommand{\emoji}[1]{}
\fi

\newcommand{\nbf}[1]{{\noindent \textbf{#1.}}}

\newcommand{\R}[1]{{%
    \textbf{%
        \ifstrequal{#1}{1}{\textcolor{red}{R#1}}{%
        \ifstrequal{#1}{2}{\textcolor{blue}{R#1}}{%
        \ifstrequal{#1}{3}{\textcolor{magenta}{R#1}}{%
        \ifstrequal{#1}{4}{\textcolor{teal}{R#1}}{%
                           \textcolor{cyan}{R#1}%
        }}}}%
    }%
}}

\definecolor{tabfirst}{rgb}{0.7, 1.0, 0.7} 
\definecolor{tabsecond}{rgb}{1, 1, 0.7} 
\definecolor{tabthird}{rgb}{1, 0.85, 0.7} 

\newcommand{\raisemath}[1]{\mathpalette{\raisemath{#1}}}

\newcommand{\Rb}{\mathbb{R}}

\definecolor{RebuttalColor}{RGB}{0,0,0}

\DeclareMathOperator*{\argmin}{arg\,min}

\definecolor{pmcolor}{RGB}{70,70,70}

%% file: sections/introduction.tex
\section{Introduction}
\label{sec:intro}
Human demonstrations are the most scalable source of data for dexterous manipulation, yet nearly every way we collect them today, whether teleoperation~\cite{aloha, wu2024gello, opentelevision}, wearable exoskeletons~\cite{umi, doglove, dexumi}, or kinesthetic teaching~\cite{sharma2025selfsupervised, chen2025dexforce}, forces humans to act through a robot embodiment, limiting dexterity to what the interface allows. Retargeting natural human motion removes this bottleneck. However, existing methods~\cite{dexpilot, anyteleop} treat retargeting as a kinematic problem of mapping fingertip positions or joint angles. Further, retargeting often requires scene reconstruction followed by heavy optimization in simulation, and even recent contact-centric formulations~\cite{contactretarget, contactanchored, hoiretarget} still presuppose an object model or a simulator. For manipulation with contact, reconstructing kinematics alone is not enough: millimeters of embodiment mismatch decide whether a replayed trajectory hovers, slips over, or crushes the object.



We present \ours, a retargeting framework that treats \fca as a first-class citizen. Instead of relying on object geometry, \ours uses per-link \emph{contact force targets} extracted from a human demonstration, comprising a center of pressure, a push direction, and a force, to solve for robot configurations that place semantically matched robot links on every contact force target inside the demonstrated friction cone. \ours produces contact force targets alongside the joint trajectory without relying on object information, therefore it needs no simulator, object mesh, or object pose estimate. Consequently, rigid, articulated, and deformable objects need no special handling.


\ours is agnostic to the demonstration source: any modality containing hand pose and contact can be used. To make heterogeneous sources interchangeable, we introduce a \emph{unified hand--contact representation} that turns monocular video (through Vid-HACO, our video contact model), tactile gloves, and marker-based motion capture into a common interface: a MANO mesh sequence with one \emph{contact force target} per hand region. Because every contact force target is bound to the human hand surface, retargeting to a new robot requires only a semantic map from hand regions to robot bodies.

On hardware, \ours pairs retargeting with a \emph{tactile tracking controller}, an integral admittance loop over joint impedance that maintains the demonstrated contact forces during replay. Together they enable \emph{open-loop replay} of dexterous human demonstrations on multiple robot embodiments mounted on a 7-DoF arm. Here \emph{open-loop} means no object pose estimate and no learned policy. The only feedback is the fingertip force loop that maintains the demonstrated contact forces. We prioritize open-loop replay to demonstrate the retargeting quality of \ours over policy learning, since a closed-loop policy confounds retargeting fidelity with other issues such as learning error, partial observability and the like. 

Concretely, our contributions are:
\begin{itemize}
    \item \textbf{Contact-Anchored Retargeting:} A simulator and object-free formulation that optimizes contact reach, friction-cone alignment, force weighting, and contact stability per frame with a fast GPU sampling optimizer.
    \item \textbf{Unified Hand--Contact Representation:} A shared interface that maps video (through Vid-HACO, our video hand-contact model), tactile gloves, and motion capture into contact force targets with whole-clip force synthesis for sources that measure no force.
    \item \textbf{Tactile Tracking Controller:} A real-world force tracking controller with feedforward force that replays contact-rich tasks open-loop faithfully on physical hardware.
    \item \textbf{Cross-Embodiment Generalization:} Cross-embodiment retargeting through a semantic map, shown on four robot hands from a 22-DoF dexterous hand down to a parallel-jaw gripper.
\end{itemize}

Across video, glove, and motion-capture sources, \ours achieves state-of-the-art contact fidelity, matching demonstrated contact on 43--78\% of contact frames where kinematic retargeting reaches under 10\%, while retargeting methods needing object information~\cite{regrind} only reach 22.5\%. Further, our trajectories transfer to the real world through open-loop replay, where the tactile tracker raises success from 4 to 25 of 30 trials over position-only replay.



%% file: sections/related_work.tex
\section{Related Work}
\label{sec:related}

\nbf{Kinematic retargeting}
DexPilot~\citep{dexpilot} established the \emph{fingertip-vector cost} used by most dexterous teleoperation systems since, including AnyTeleop and \texttt{dex-retargeting}~\citep{anyteleop}, learned pose imitators~\citep{telekinesis, geort}, and VR and exoskeleton interfaces~\citep{opentelevision, bunnyvisionpro, ace, doglove}. All these approaches retarget hand configuration to hand configuration. Two recent geometric retargeting methods, TopoRetarget~\citep{toporetarget} and Morphometric Imitation~\citep{morphometric}, go beyond fingertip matching, although their optimization still relies on an object reference. Neither had released code at the time of submission, so we cannot compare against them. \ours keeps the optimization view but optimizes contact locations, push directions, and forces.

\nbf{Contact-based retargeting}
Contact Transfer~\citep{lakshmipathy2022contacttransfer} and its sequel~\citep{contactretarget} transfer contact areas instead of joint angles. Concurrent work anchors retargeting on contact structure, under the same name~\citep{contactanchored} or in the object frame~\citep{hoiretarget}, or on demonstrated forces~\citep{reforce, atar2026contact}. Each needs an object model, a simulator, or a learned tracker. \ours needs none, and adds a push direction and a force to every contact.

\nbf{Dynamic retargeting}
A complementary line transfers demonstrations through simulation with the object in the loop~\citep{maniptrans, dextrack, quasisim, pgdm, objdex, omnigrasp, dexmachina}, optimizes robot and object trajectories jointly~\citep{omniretarget}, or rewards contact wrenches computed on the object~\citep{chord}. These methods capture contact dynamics in simulation but need an object model and a simulator per task. \ours needs neither, since its contact force targets are obtained directly in the real world.

\nbf{Learning from human data}
Many pipelines recover hand and object pose from video or wearable capture devices and retarget kinematically for imitation or RL~\citep{dexmv, dexvip, videodex, doasido, dexcap, h2o, omnih2o, humanplus}. Concurrent ones preserve contact, but anchored on a reconstructed object~\citep{sun2026egocpt, ren2026c2dex, mao2026omnihoi, niu2026egoaero}. Hardware interfaces~\citep{aloha, umi, airexo, dexumi} make collection cheaper but tie the human to a device built per robot. Human contact and force encode grasp function~\citep{contactpose, grab, tactileglove, contactgrasp} and improve contact-rich learning~\citep{forcemimic, acp, forcesight, ftf}, yet retargeting rarely preserves them, which is the gap \ours closes.

\nbf{Dexterous hand controllers}
Tactile and force feedback improve multifingered control, from contact-event grasp controllers~\citep{romano2011} and tactile-driven policies~\citep{tdex, vitac3d} to tactile servoing and grasp-force regulation~\citep{tactileservo, graspadapt, slipgrip}. These controllers regulate forces chosen by a policy or a heuristic. \ours outputs tactile targets from a demonstration, so a robot hand can replay it open-loop.

\nbf{Contact and force from vision}
Dense hand contact has been predicted from hand--object geometry~\citep{grady2021contactopt} and from images~\citep{haco}, and concurrently from video together with pressure~\citep{jeon2026hope, moon2026rlhnd} or force~\citep{ilic2026egophi}. Forces have also been fitted to the tracked object's wrench~\citep{pham2015force, pham2018}. \ours consumes any contact source and synthesizes forces from grasp balance without an object model.


%% file: sections/method.tex
\section{Method}
\label{sec:method}

\begin{figure*}[t]
    \centering
    \includegraphics[width=0.8\linewidth]{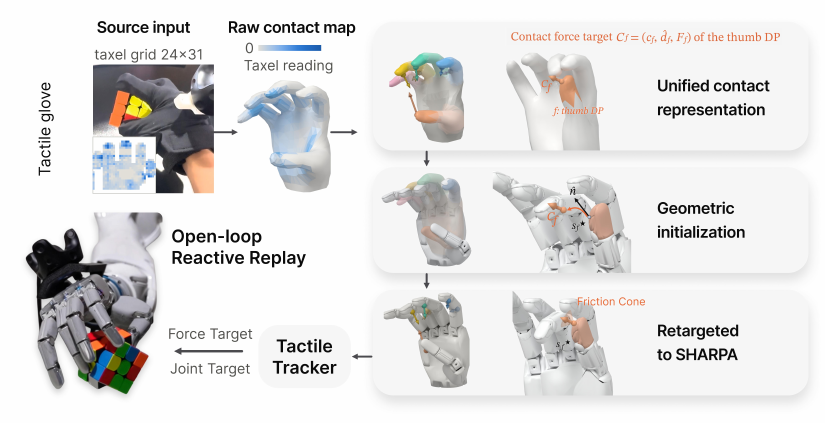}
    \caption{\textbf{\ours overview.} Video, glove, and motion capture are converted into a unified hand--contact representation, a MANO mesh with one contact force target per semantic link (\cref{sec:method:uhcr}). Geometric initialization maps human joint angles to the robot skeleton, followed by contact-anchored retargeting that aligns the corresponding robot pads with the contact force targets (\cref{sec:method:retarget}). Joint and contact force targets are replayed by a tactile tracking controller (\cref{sec:method:replay}) on hardware.}
    \label{fig:pipeline}
\end{figure*}

Given a demonstration with hand pose and contact, \ours (\cref{fig:pipeline}) produces a robot trajectory that reproduces its \fca, not only its kinematics. Per frame we solve for the hand pose $q = (t, R, \theta)
\in \mathbb{R}^3 \times \mathrm{SO}(3) \times \mathbb{R}^{N}$: wrist position, wrist rotation as an axis--angle vector, and the $N$ joint angles ($N{=}22$ for the SHARPA Wave), without a physics simulator or object model as input. Every \emph{contact force target} is bound to the hand surface, so deformable and articulated objects can be handled without additional burden. Hats ($\hat{\phantom{s}}$) denote unit vectors, $q^{\star}$ denotes the optimal solution of a frame, $q^{-}$ denotes the previous frame's solution.
\subsection{Unified Hand--Contact Representation}
\label{sec:method:uhcr}

Every data source is reduced to three components (\cref{fig:regions}): a MANO~\citep{mano} hand mesh sequence, a vocabulary of semantic links (3 regions per finger, 15 finger regions plus the palm region, see \cref{app:details} for details) shared by every source and every robot, and at most one \emph{contact force target} per link and frame, denoted by $C_f = (c_f, \hat{d}_f, F_f)$, where $f$ is the finger link and $\mathcal{F}$ the full robot link set. The center of pressure $c_f \in \Rb^3$ is constrained onto link surface $f$, $\hat{d}_f$ denotes a unit push direction into the object, and $F_f \geq 0$ denotes the force magnitude. Video does not provide contact force, so we synthesize it from the contact predictions. Source-specific processing into the shared representation is detailed in \cref{app:data_unification}.

\subsection{Vid-HACO: Contact from Monocular Video}
\label{sec:method:vidhaco}

For monocular video, we introduce Vid-HACO, a video-based hand–object contact estimation model that predicts contact probabilities for all 778 MANO vertices, enabling video demonstrations to be converted into our unified hand–contact representation. Given an 8-frame clip ending at frame $t$, Vid-HACO uses hand crops obtained from HaWoR~\citep{hawor}, a world-frame extension of HaMeR~\citep{hamer}. A video-pretrained V-JEPA 2.1 ViT-L encoder \citep{vjepa21} and a DPT decoder \citep{dpt} extract multi-scale features. Two prediction heads estimate vertex-level contact logits and a frame-level binary contact indicator, respectively. Temporal context helps distinguish contact transitions during approach, grasp, and release.

Following HACO~\citep{haco}, we use a 10\,mm distance threshold to label vertices as contacting the touched surface and set the probability threshold to 0.52 for contact prediction. We train on 1.55M frames from 12 datasets spanning egocentric, third-person, and web views, holding out 10\% of each dataset for validation. We fine-tune the encoder end to end with HACO's class-balanced vertex loss~\citep{cbloss} and an additional binary cross-entropy loss on the frame head, retaining contact-free frames during training. Dataset and implementation details appear in \cref{app:vidhaco}.

\subsection{Contact-Anchored Retargeting}
\label{sec:method:retarget}

\nbf{Geometric initialization} We first retarget human joint angles directly to the robot skeleton using closed form approaches (i.e., without optimization) and adjust the wrist pose to align the robot and human fingertips in 3D. Because the thumb joint ranges and motion axes cannot be mapped directly, we compute its pose using a fixed affine transformation derived from its bone directions and pad normal (\cref{app:init}). The resulting configuration $q^{\mathrm{init}}$ is deterministic and warm starts the following optimization (\cref{fig:init}).


\nbf{Problem formulation} For each frame, we optimize the robot joint configuration $q^\star$ by solving:
\begin{align}
\label{eq:objective}
q^{\star} = \arg\min_{q \in \mathcal{Q}} \; &E_{\mathrm{contact}}(q) + E_{\mathrm{style}}(q) + E_{\mathrm{smooth}}(q)\nonumber\\ + &E_{\mathrm{collision}}(q),
\end{align}
where $E_{\mathrm{contact}}$ aligns the robot's fingertip pads with \fca, $E_{\mathrm{style}}$ encourages unconstrained (free) fingers to mimic human poses, $E_{\mathrm{smooth}}$ maintains smooth joint movements, while $E_{\mathrm{collision}}$ prevents robot self-collisions and object penetration. The feasible set $\mathcal{Q}$ enforces joint position and velocity limits across all joints. A single set of loss weights $w$ is used across all motion sequences. The weights appear inside each energy term below and are listed in \cref{tab:solver} of \cref{app:details}.

\nbf{Contact group} Each robot contact region $f \in \mathcal{F}$ on the pad surface $\mathcal{S}_f(q)$ is defined by its 3D position $s_f(q)$ and outward surface normal $\hat{n}(s)$, which specifies the central axis of its friction cone. To prevent physically unrealistic assignments (e.g., satisfying a fingertip contact using a mid-link surface), we select an optimal contact point $s^{*}_f(q)$ by balancing geometric contact proximity against human contact preferences:
\begingroup\setlength{\abovedisplayskip}{3pt}\setlength{\belowdisplayskip}{3pt}\setlength{\abovedisplayshortskip}{1pt}\setlength{\belowdisplayshortskip}{1pt}
\begin{equation}\label{eq:selector}s^*_f(q) = \argmin\nolimits_{s \in \mathcal{S}_f(q)} \|s - c_f\|^2 + \alpha_f \|s - T_f\psi_f\|^2.\end{equation}\endgroup
Here, $c_f$ is the target contact location and $\psi_f$ is the human contact preference remapped onto the robot link: the point on the link's palmar centerline at the same fractional position $u_f$ at which the human touched the corresponding finger segment, $u_f = \mathrm{clip}\big(\tfrac{(c_f - j_0)\cdot(j_1 - j_0)}{\|j_1 - j_0\|^2},\, 0,\, 1\big)$, where $j_0$ and $j_1$ are the joints bounding that segment (e.g., DIP and fingertip for the distal region). The link pose $T_f(q)$ maps $\psi_f$ into the world frame. Because $s$ and $\psi_f$ lie on the same rigid link, the second term acts as a fixed offset in the link frame that guides the selection of $s^{*}_f(q)$. Once selected, $s^{*}_f(q)$ is evaluated in the total contact objective $E_{\mathrm{contact}}(q)$:

\begin{equation}
\label{eq:contact}
\begin{split}
E_{\mathrm{contact}}(q) = {}& \sum_{f \in \mathcal{F}} \rho_f^2 \Big[\, w_{\mathrm{c}}^2 \Big( \underbrace{\big\| s^{*}_f(q) - c_f \big\|^2}_{\text{contact reach}} \\
& + \underbrace{\lambda r^2 \big(1 - \hat{n}(s^{*}_f(q)) \cdot \hat{d}_f\big)}_{\text{friction cone}} \Big) \\
& + \underbrace{w_{\mathrm{fix}}^2\, \big\| s^{*}_f(q) - \Delta T_f(q,q^{-})\, s^{*}_f(q^{-}) \big\|^2}_{\text{anti-slip}} \Big]
\end{split}
\end{equation}

Here, $\rho_f = 1 + \frac{1}{2} (F_f / \max_{f'} F_{f'})$ scales contact priority by applied force, ensuring high-force contacts are prioritized. The friction cone term aligns the pad normal with the desired pushing direction $\hat{d}_f$ to prevent slipping. Its weight $\lambda r^2$ has a geometric reading: the length scale $r$ converts the misalignment $1 - \hat{n} \cdot \hat{d}_f$ into an equivalent squared position error, so that both terms share units of mm$^2$, and the dimensionless gain $\lambda$ prices direction against position. With $\lambda = 2$ and $r = 10\,\text{mm}$, a $30^\circ$ orientation error costs as much as a $5\,\text{mm}$ position deviation. Finally, the anti-slip term penalizes contact migration across the pad surface by transforming the previous contact point into the current link frame with $\Delta T_f(q,q^{-}) = T_f(q)T_f(q^{-})^{-1}$, encouraging the contact to remain at the same location on the fingerpad across frames (\cref{fig:pipeline}).

\nbf{Style, smoothness, collision}
When a finger is unconstrained by contact force targets, $E_{\mathrm{style}}$ encourages the robot hand to mimic human motion. Specifically, $E_{\mathrm{style}}$ pulls each robot pad toward its corresponding MANO fingertip, aligns the pad's belly axis with the MANO surface normal, and aligns individual finger segment orientations with the human pose. $E_{\mathrm{smooth}}$, on the other hand, applies a quadratic penalty on joint motion relative to the previous frame to maintain temporal continuity. Finally, $E_{\mathrm{collision}}$ uses a soft hinge barrier to penalize hand self-collisions and object penetrations. Inter-finger clearance uses capsule-to-capsule gaps. Object penetration queries pad samples against an object signed distance field (SDF) or, without a mesh (video, glove, GRAB), against a pseudo-object SDF built from the convex hull of the clip's contact force targets $c_f$ in the wrist frame (\cref{fig:pseudoobject}, \cref{app:details}).
\nbf{Sampling-based optimization} Because the objective in \cref{eq:objective} is non-smooth, we minimize it using zeroth-order optimization---specifically, an annealed Cross-Entropy Method (CEM)~\citep{mppi, icem} whose noise schedule follows the diffusion-style annealing of DIAL-MPC~\citep{dialmpc}. Each frame is warm-started with the geometric initialization $q^{\mathrm{init}}$. At each of the $K = 16$ iterations, $N = 1024$ candidate configurations are sampled around the mean, clipped to the feasible set $\mathcal{Q}$, and evaluated in a single parallelized pass. The mean is then updated using a softmax weighting over the top 5\% lowest-cost candidates while noise is progressively annealed. A zero-phase Savitzky--Golay filter (15 frames) smooths the result. The solver runs at 10 FPS on one RTX 5090.


\input{sections/experiment_figures}

\subsection{Replay: Tactile Tracking Controller}
\label{sec:method:replay}

\begin{figure}[t]
    \centering
    \includegraphics[width=0.85\linewidth]{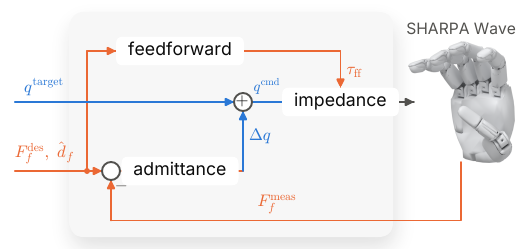}
    \caption{\textbf{Tactile tracker.} Each finger uses fingertip force feedback to track the desired contact force through an admittance loop whose curls form the joint offset $\Delta q$, while feedforward torque directly applies the target force through the joint impedance controller (\cref{app:tracker}).}

    \label{fig:tracker_diagram}
\end{figure}

The retargeting pipeline outputs a joint trajectory $q^{\mathrm{target}}$ and, per finger $f$, a desired normal force $F^{\mathrm{des}}_f$ along the push direction $\hat{d}_f$. A tactile controller tracks both (\cref{fig:tracker_diagram}). An integral-dominant admittance on the error between $F^{\mathrm{des}}_f$ and the measured fingertip normal force $F^{\mathrm{meas}}_f$ produces a scalar \emph{curl} $\kappa_f \in [0,1]$ (\cref{eq:admittance} in \cref{app:tracker}), which closes the finger along a fixed unit synergy $\hat{\psi}_f$. The summed closures form a joint offset $\Delta q = \textstyle\sum_f \kappa_f\,\hat{\psi}_f$ that updates the joint target, $q^{\mathrm{cmd}} = q^{\mathrm{target}} + \Delta q$, while a feedforward torque injects the desired force through the fingertip Jacobian $J_f$ into the joint impedance controller with gains $K_p, K_d$:
\begin{equation}
\label{eq:controller}
\begin{aligned}
\tau &= K_p\,(q^{\mathrm{cmd}} - q) + K_d\,(\dot{q}^{\mathrm{cmd}} - \dot{q}) + \underbrace{k_{\mathrm{ff}}\, J_f^{\top} F^{\mathrm{des}}_f\, \hat{d}_f}_{\tau_{\mathrm{ff}}}.
\end{aligned}
\end{equation}
Because the joints are compliant, the curl adds a bounded squeeze rather than a rigid displacement. A contact state machine freezes the integrator on sensor dropouts and arms the feedforward (gains in \cref{app:tracker}).


\subsection{Cross-Embodiment Transfer}
\label{sec:method:embodiment}

Because \fca are parameterized in the unified representation, transfer to a new embodiment needs only a \emph{semantic map} (\cref{fig:semantic_map}, \cref{app:details}). Defined anchors, cost groups, and weights can be used across embodiments. Five-fingered hands map one-to-one, region to link. Hands with fewer fingers map many-to-one, for instance, the Unitree Dex3-1 receives the middle, ring, and pinky regions on its third finger, with simultaneous anchors merged into a force-weighted centroid, and a parallel-jaw gripper maps the thumb to one jaw and the four fingers to the other. We show four embodiments' semantic mapping in \cref{fig:real,fig:embodiments}.


%% file: sections/experiment_figures.tex
\begin{figure}[t]
    \centering
    \includegraphics[width=\linewidth]{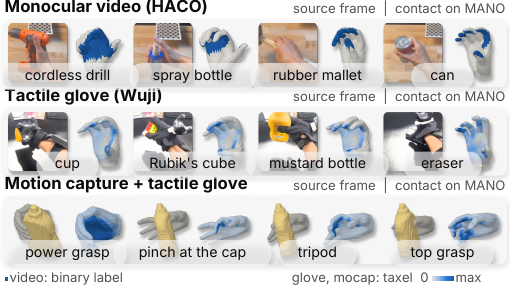}
    \caption{\textbf{Demonstration sources.} We show some demonstration exmaples from monocular video (Vid-HACO contact), tactile glove (taxel pressure), and motion capture with the tactile glove  (taxel pressure).
}
    \label{fig:gallery}
\end{figure}

\begin{figure}[t]
    \centering
    \includegraphics[width=\linewidth]{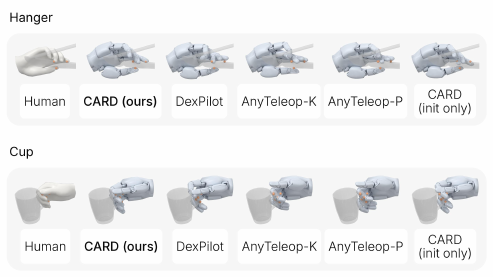}
    \caption{\textbf{Contact-anchored vs.\ kinematic retargeting} on two video clips. Per clip: the human source, \ours, DexPilot, AnyTeleop-K, AnyTeleop-P, and \ours's geometric initialization (init only), the columns of \cref{tab:contact_quality}. The scanned object is for visualization only, and orange spheres are the \fcas. Kinematic baselines reach the object but do not track the contact.}
    \label{fig:baselines}
\end{figure}

%% file: sections/experiments.tex
\begin{table*}[t]
    \centering
    \footnotesize
    \setlength{\tabcolsep}{5pt}
    \renewcommand{\arraystretch}{0.85}
    \caption{\textbf{Contact fidelity} on our data (video, glove, motion capture) and on GRAB. Rates in percent, pooled over anchor-frames. Millimeter columns are medians. \textbf{Bold}: best per bed and metric. \underline{Underline}: best published baseline. On GRAB we additionally compare against ReGrind ported to SHARPA Wave, which takes the GRAB mesh as input. \ours uses only the contact-reconstructed pseudo-object (\cref{fig:pseudoobject}), and the GRAB mesh only to score penetration. The glove$+$motion capture gap combines the two beds' medians weighted by anchor-frames.}
    \label{tab:contact_quality}
            \begin{tabular}{lcccccc}
        \toprule
        & \ours & DexPilot & AnyTeleop-K & AnyTeleop-P  & ReGrind & \ours \\
         & & \citep{dexpilot} & \citep{anyteleop} & \citep{anyteleop} & \citep{regrind} (w/ Object)& (init only) \\
        \midrule
        \multicolumn{7}{l}{\emph{Our data: monocular video (20 clips)}} \\
        \quad Contact rate (\%) $\uparrow$& \textbf{61.4} & \underline{9.7} & 8.3 & 7.6 & -- & 20.5 \\
        \quad Correct region (\%) $\uparrow$& \textbf{60.1} & \underline{9.5} & 8.0 & 7.5 & -- & 19.2 \\
        \quad In-cone (\%) $\uparrow$& \textbf{35.2} & 4.2 & 4.7 & \underline{5.0} & -- & -- \\
        \quad Gap (mm) $\downarrow$& \textbf{2.40} & \underline{9.99} & 10.80 & 13.19 & -- & 5.76 \\
        \midrule
        \multicolumn{7}{l}{\emph{Our data: tactile glove $+$ motion capture (66 clips)}} \\
        \quad Contact rate (\%) $\uparrow$& \textbf{77.7} & \underline{3.8} & 3.4 & 3.5 & -- & 15.9 \\
        \quad Correct region (\%) $\uparrow$& \textbf{76.9} & \underline{3.7} & 3.3 & 3.2 & -- & 15.8 \\
        \quad In-cone (\%) $\uparrow$& \textbf{46.0} & 1.1 & 1.0 & \underline{2.3} & -- & -- \\
        \quad Gap (mm) $\downarrow$& \textbf{1.38} & 17.17 & \underline{16.67} & 23.22 & -- & 9.32 \\
        \midrule
        \multicolumn{7}{l}{\emph{GRAB (100 clips)}~\citep{grab}} \\
        \quad Contact rate (\%) $\uparrow$& \textbf{43.1} & 4.9 & 4.9 &  4.6 &\underline{22.5}& 19.6 \\
        \quad Correct region (\%) $\uparrow$& \textbf{40.6} &  4.8 & 4.9 &  4.5 & \underline{20.7} & 17.8 \\
        \quad In-cone (\%) $\uparrow$& \textbf{16.2} & 1.7 & 1.7 &1.7 &\underline{5.2}&  3.8 \\
        \quad Gap (mm) $\downarrow$& \textbf{3.56} & 20.90 & 20.45 & 20.66 &  \underline{7.86}  &13.78 \\
        \quad Penetration (mm) $\downarrow$& \textbf{7.96} & 12.71 & 12.79 & 13.02 & \underline{9.80}& 11.15 \\
        \bottomrule
        \end{tabular}
\end{table*}

\begin{figure}[t]
    \centering
    \includegraphics[width=0.8\linewidth]{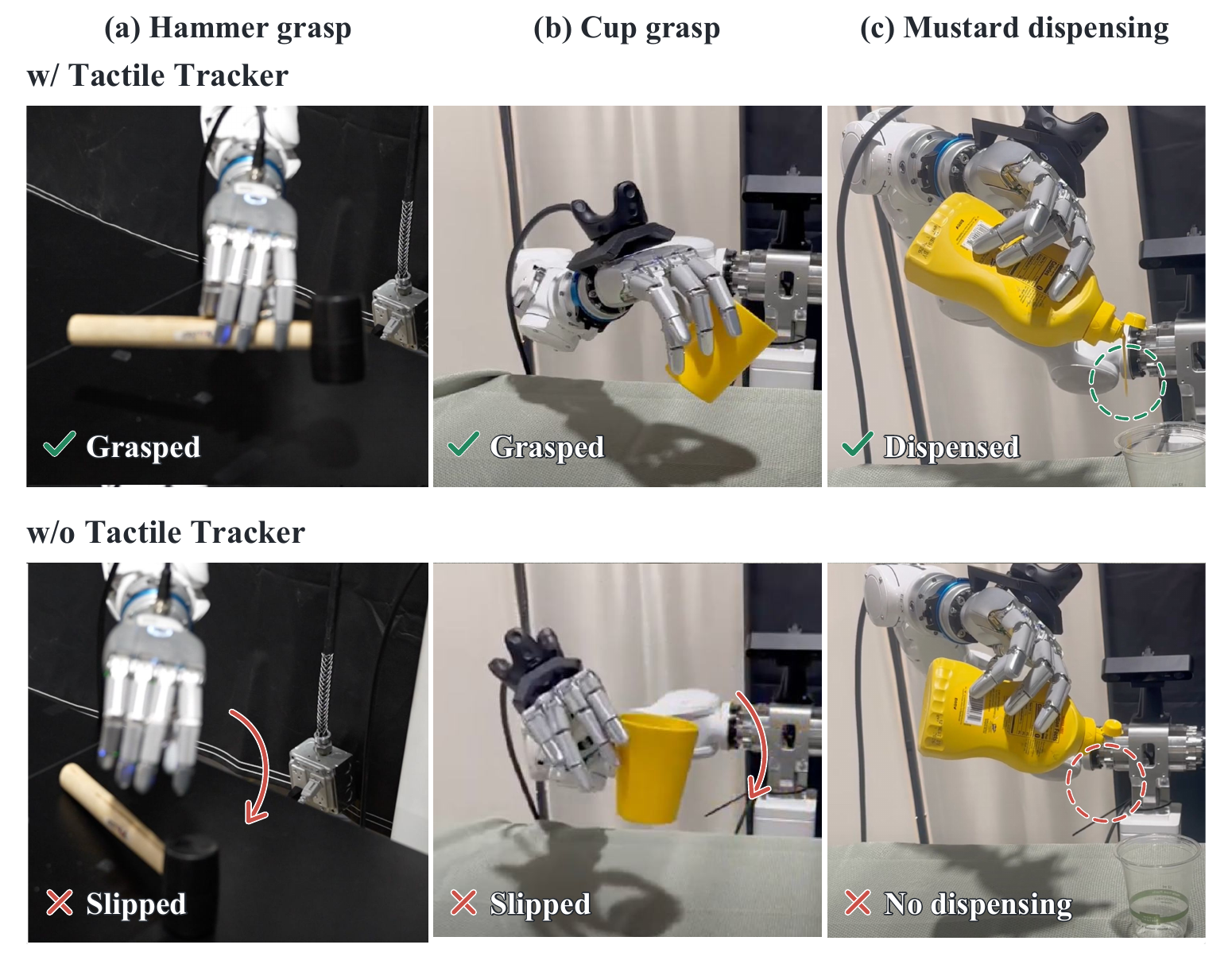}
    \caption{\textbf{Tactile tracker ablation.} We show the same trajectories replayed with (top) and without (bottom) tactile tracking. Tactile tracking maintains the desired contact forces, enabling stable grasps and successful task execution.}

    \label{fig:tactile_ablation}
\end{figure}

\begin{figure*}[t]
    \centering
    \includegraphics[width=0.9\linewidth]{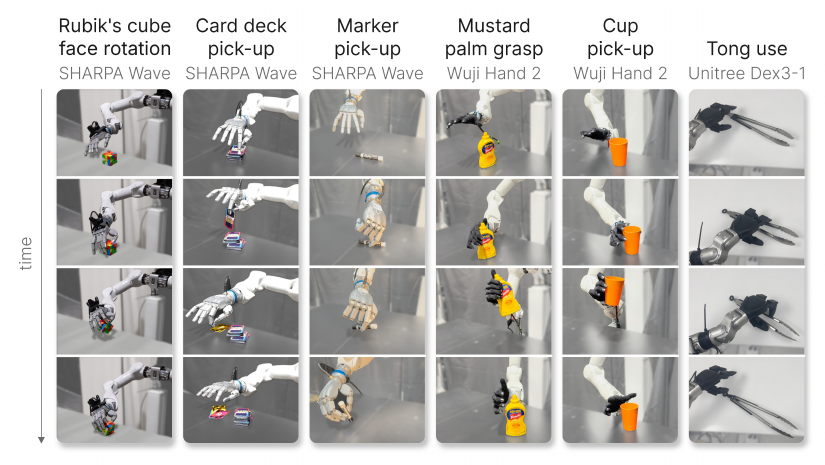}
    \caption{\textbf{Replay on hardware.} We show a series of snapshots of \ours trajectories replayed on the SHARPA Wave hand, Wuji Hand~2, and Unitree Dex3-1.  The articulated Rubik's cube and the bending deck of cards use the same pipeline as the rigid objects.}
    \label{fig:real}
\end{figure*}

\begin{figure}[t]
    \centering
    \includegraphics[width=0.8\linewidth]{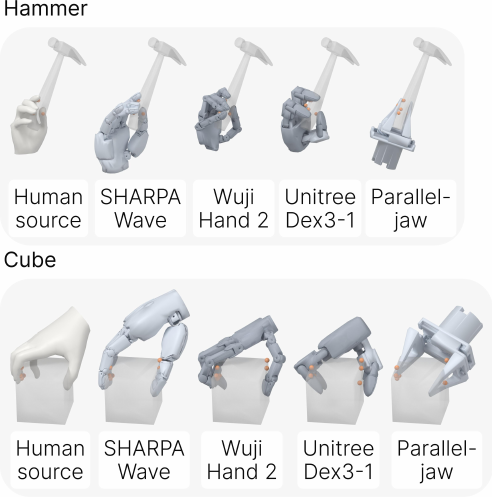}
    \caption{\textbf{Cross-embodiment transfer.} Two GRAB demonstrations (rows) retargeted to SHARPA Wave, Wuji Hand~2, Unitree Dex3-1, and a parallel-jaw gripper (columns, after the human source). Only the semantic map changes. Orange spheres are the contact force targets.}
    \label{fig:embodiments}
\end{figure}

\begin{table}[t]
    \centering
    \caption{\textbf{Contact from monocular video.} Vid-HACO versus released HACO~\citep{haco} and HACO-R (retrained on our data), on 12-dataset validation and held-out HRDexDB~\citep{hrdexdb}. Percent, threshold 0.52 except for the threshold-free AP. Vertex metrics are pooled over all vertices and frames. Frame F1 measures whether any vertex is in contact.}
    \label{tab:contact}
    \small
    \setlength{\tabcolsep}{4.5pt}
    \begin{tabular}{@{}lccccc@{}}
    \toprule
    & \multicolumn{4}{c}{Vertex} & Frame \\
    \cmidrule(lr){2-5}
    & F1 $\uparrow$ & Prec.\ $\uparrow$ & Rec.\ $\uparrow$ & AP $\uparrow$ & F1 $\uparrow$ \\
    \midrule
    \multicolumn{6}{@{}l}{\emph{12-dataset validation}} \\
    \quad HACO & 57.1 & 44.4 & 80.0 & 51.7 & 86.1 \\
    \quad HACO-R & 74.5 & 63.4 & \textbf{90.3} & 70.4 & 86.1 \\
    \quad Vid-HACO & \textbf{84.3} & \textbf{80.4} & 88.5 & \textbf{92.4} & \textbf{98.3} \\
    \midrule
    \multicolumn{6}{@{}l}{\emph{HRDexDB (held out)}} \\
    \quad HACO & 63.3 & 56.5 & \textbf{71.9} & 54.3 & 76.3 \\
    \quad HACO-R & 62.6 & 57.1 & 69.2 & 54.5 & 75.9 \\
    \quad Vid-HACO & \textbf{67.2} & \textbf{67.9} & 66.5 & \textbf{72.2} & \textbf{95.3} \\
    \bottomrule
    \end{tabular}\vspace{-1em}
\end{table}

\section{Experiments}
\label{sec:exp}

We evaluate \ours across five aspects: contact fidelity against kinematic baselines (\cref{sec:exp:setup}), force preservation in open-loop hardware replay (\cref{sec:exp:real}), transfer across embodiments (\cref{sec:exp:embodiment}), contact estimation from video (\cref{sec:exp:vidhaco}), and the performance of a physics-based retargeting variant, \ours--physics (\cref{sec:exp:physics}). We also provide additional ablations and contact analyses: \cref{app:ablation,app:by_source,app:contact}.

\subsection{Contact fidelity against kinematic retargeting}
\label{sec:exp:setup}

\nbf{Data} We evaluate on two datasets: clips captured in house (\cref{fig:gallery}: 20 casual phone videos, 18 tactile glove with VIVE tracker clips, and 48 motion-capture with tactile glove clips) and 100 public GRAB clips~\citep{grab} with ground-truth object meshes. Every method is retargeted to the 22-DoF SHARPA Wave.

\nbf{Metrics} A contact is considered \emph{established} when a \fcas lies within 2\,mm of the assigned link surface and the nearest pad sample is on the palmar face. \emph{Contact rate} measures the fraction of anchor-frames with established contact. \emph{Correct-region rate} further requires contact on the correct region according to the demonstration, while \emph{in-cone rate} additionally requires the pad normal to align within $45^\circ$ of the demonstrated push direction. \emph{Gap} is the median distance from the contact force target to the pad and \emph{penetration} the median deepest incursion into the object (\cref{app:metrics}).

\nbf{Baselines} We compare against the kinematic retargeting baselines: DexPilot~\citep{dexpilot} and the two objectives of AnyTeleop~\citep{anyteleop}, key-vector (AnyTeleop-K) and fingertip-position (AnyTeleop-P). We also report ReGrind~\citep{regrind} (w/ Object), a contact-preserving retargeting method that requires the object mesh, and \ours's geometric initialization (init only). TopoRetarget~\citep{toporetarget} and Morphometric Imitation~\citep{morphometric} have not released code at submission time and are therefore not compared.

\nbf{Results} \Cref{tab:contact_quality} compares contact fidelity across both datasets. Kinematic baselines establish the demonstrated contact on only 3--10\% of anchor-frames, with a median gap of 10--23\,mm. In contrast, \ours establishes contact on 61\% of anchor-frames on video, 78\% on the glove and motion-capture clips, and 43\% on GRAB, while reducing the gap to 1.4--3.6\,mm. Importantly, nearly all established contacts occur on the correct finger, and over half of anchor-frames also satisfy the in-cone constraint. On GRAB, where ground-truth object meshes allow us to measure penetration, \ours reaches 43\% contact against 22.5\% for ReGrind, which takes the object mesh as input, and also produces the least object penetration (7.96\,mm) (\cref{fig:baselines}). We also find that contact fidelity correlates with contact signal quality: \ours can achieve 89\% contact on the tactile glove clips, where contact is measured, 76\% on motion capture, 61\% on video, where it is inferred, and 43\% on GRAB, where contact is calculated from geometric proximity (\cref{app:by_source}).

\subsection{Open-loop replay on hardware}
\label{sec:exp:real}

To directly evaluate the fidelity of \ours's retargeted actions, we replay the trajectories open-loop on real hardware, eliminating confounding factors such as policy learning. Under this setting, task success depends directly on the faithful reproduction of task-critical contacts and forces. \cref{fig:real} shows sequential snapshots of our hardware replays, including picking up a deformable card from a deck and rotating an articulated Rubik’s Cube with a single finger. All tasks use our tactile tracking controller to preserve contact forces. Full videos are available on our project website.

\nbf{Tactile tracker ablation} We replay identical \ours trajectories on the SHARPA Wave with and without the tactile tracker on three real-world tasks, 10 trials each (\cref{fig:tactile_ablation}). Position-only replay succeeds on 1/10 hammer lifts, 3/10 cup lifts, and 0/10 mustard dispenses: the hammer and cup slip out during the lift while the bottle is lifted but not squeezed. With the tactile tracker the same trajectories succeed on 8/10, 10/10, and 7/10 respectively. Reproducing the joint motion alone does not reproduce the intended contact forces.

\subsection{Cross-embodiment transfer}
\label{sec:exp:embodiment}

Here we demonstrate that the same \fca formulation can retarget to new embodiments such as the Wuji Hand~2, the Unitree Dex3-1 and a parallel-jaw gripper (\cref{fig:embodiments}) by only changing the semantic mapping. Specifically, we show hardware rollouts with the first two embodiments in \cref{fig:real}. Unitree Dex3-1 has no tactile sensing, so we scale its grasp actions 1.2$\times$. Although the Wuji Hand~2 also has no tactile sensing, we found that it needs no scaling, since its compliant inner surfaces conform to the object.

\subsection{Contact estimation from video}
\label{sec:exp:vidhaco}

\nbf{Setup} We compare our video contact model Vid-HACO with HACO~\citep{haco} and HACO-R, HACO retrained on the same 12-dataset training set as ours. We evaluate on the $\sim$173k-frame validation split and held-out HRDexDB~\citep{hrdexdb}: 434 egocentric videos of 94 objects (111.9k frames). All models use identical hand crops, the 10\,mm contact rule, and threshold 0.52.

\nbf{Results} \Cref{tab:contact} shows that Vid-HACO achieves 84.3 vertex F1 on validation and 67.2 on HRDexDB, improving held-out F1 by 3.9 points over released HACO. Retraining HACO improves validation F1 to 74.5 but leaves held-out F1 at 62.6. Vid-HACO exceeds this matched-data baseline by 4.6 F1 and 17.7 AP. Frame F1 reaches 95.3 versus 75.9 for both baselines, which predict contact on every frame. This is important for retargeting since hallucinated contacts during a reach can pull robot fingers toward empty space. Further analysis is provided in \cref{app:contact}.

\subsection{\ours with physically plausible retargeting}
\label{sec:exp:physics}

\begin{table}[t]
    \centering
    \small
    \setlength{\tabcolsep}{8pt}
    \caption{\textbf{\ours-physics.} We compare \ours-physics, our physics retargeting variant, against SPIDER's physics stage~\citep{spider}. Strict task success in percent on 48 motion-capture demonstrations (2 seeds) and 50 GRAB clips (1 seed). A task counts as solved when at least one seed succeeds (\cref{app:physics}).}
    \label{tab:physics}
    \begin{tabular}{lcc}
        \toprule
        & Motion capture (48) & GRAB (50) \\
        \midrule
        SPIDER~\citep{spider} & 40\% & 12\% \\
        \ours-physics & \textbf{52\%} & \textbf{32\%} \\
        \bottomrule
    \end{tabular}\vspace{-1em}
\end{table}

We further test whether \fca and the contact objective improve physics-based retargeting. We build \ours-physics on SPIDER~\citep{spider}, a simulation-based retargeting framework that tracks kinematic references using a sampling-based optimization solver. \ours-physics modifies SPIDER in three ways: (1) initialize from \ours's retargeted trajectories, (2) replace SPIDER's contact-guidance term with our contact objective $E_{\mathrm{contact}}(q)$, and (3) extract \fca using our pipeline (details and weights in \cref{app:physics}).

We evaluate on 50 GRAB and 48 motion-capture trajectories with ground-truth contact, object meshes, and object trajectories. \ours-physics raises strict task success from 12\% to 32\% on GRAB and from 40\% to 52\% on motion capture (\cref{tab:physics}), showing evidence that our contact representation also benefits physics-based retargeting.

%% file: sections/conclusion.tex

\section{Conclusion}
\label{sec:conclusion}

We presented CARD, a simulator- and object-free retargeting framework that preserves task-critical contacts and forces across diverse human demonstration sources and robot embodiments. Combined with a tactile tracking controller, CARD enables faithful open-loop hardware replay across rigid, articulated, and deformable objects. Our results demonstrate substantial gains in contact fidelity and task success, establishing contact preservation as a scalable interface for transferring human dexterity to robots.

\nbf{Limitations and future work} Retargeting quality is bounded by the input demonstration, and open-loop replay cannot recover from a large object displacement. We will use \ours-generated data to train closed-loop sensorimotor policies that close this gap, and to scale human demonstrations into contact-preserving robot data.

\section{Acknowledgement}
The authors would like to thank Chaoyi Pan, Tingfan Wu, and Bipasha Sen for their insightful discussions and valuable feedback, and Xinyao Li and Bruno Chen for their assistance with the experiments. We also gratefully acknowledge the Amazon FAR group for their continued support throughout this work.

\label{sec:conclusion}

%% file: sections/appendix.tex
\section{Unifying Vision, Glove, and Motion-Capture Demonstrations}
\label{app:data_unification}

\nbf{Data sources}
\emph{Vision:} A monocular phone video is tracked in the world frame with HaWoR~\citep{hawor}, while per-vertex hand--object contact is predicted by Vid-HACO (\cref{sec:method:vidhaco,sec:exp:vidhaco}). Hand segmentation and metric scale are obtained using off-the-shelf models~\citep{sam2, depthanything}.
\emph{Tactile glove:} A 744-taxel pressure grid provides measured contact and pressure, with each taxel registered to a personalized MANO mesh, and a wrist-mounted tracker provides hand motion.
\emph{Motion capture with tactile glove:} The same tactile glove is worn under OptiTrack markers, so the glove supplies the contact and pressure while the markers provide drift-free wrist motion, together with a tracked object mesh used only for evaluation. GRAB~\citep{grab} and ARCTIC~\citep{arctic} are incorporated through their native annotations and loaders.

\nbf{From sensor measurements to \fca}
All three sources are converted into the same \fca representation. Each source delivers, per frame and link, a set of contacting MANO vertices $v$ with a weight $w_v$: the registered taxel pressure for the glove and motion-capture sources, and $w_v = 1$ for video, where Vid-HACO outputs a binary contact mask, so every contacting vertex counts equally. GRAB provides contact from hand--object proximity in place of pressure. The center of pressure $c_f$ follows in three steps. (i)~The contacting vertices are expressed in the link's 2-D pad chart, with the bone as the $x$ axis and the pad normal as $z$, and reduced to the weighted centroid $(\bar{x}_f, \bar{y}_f) = \sum_v w_v (x_v, y_v) / \sum_v w_v$. (ii)~The $(\bar{x}_f, \bar{y}_f)$ track is smoothed over time with a centered 11-frame moving average, which adds no lag and, because it acts on the chart, keeps the point on the pad. (iii)~The smoothed point is mapped back to 3-D and snapped onto the MANO surface, giving $c_f$. The contact direction $\hat{d}_f$ is the weighted mean of the contacting vertices' outward surface normals, oriented toward the object, and the raw force magnitude is the total weight $\sum_v w_v$ of the patch. Because video and GRAB do not measure force, we estimate their forces using a whole-clip grasp-balance procedure that finds the smallest smooth force correction driving the net wrench of each grasp toward zero. The same balance principle underlies force estimation from video, where contact forces are fitted to the tracked object's Newton--Euler equations~\citep{pham2015force, pham2018}; without an object model, we instead balance the net wrench of the contacts themselves. Finally, short contact blinks are removed, contact onsets within each grasp are synchronized, and video-derived contacts are given a short temporal pre-roll.

\begin{figure}[t]
    \centering
    \includegraphics[width=\linewidth]{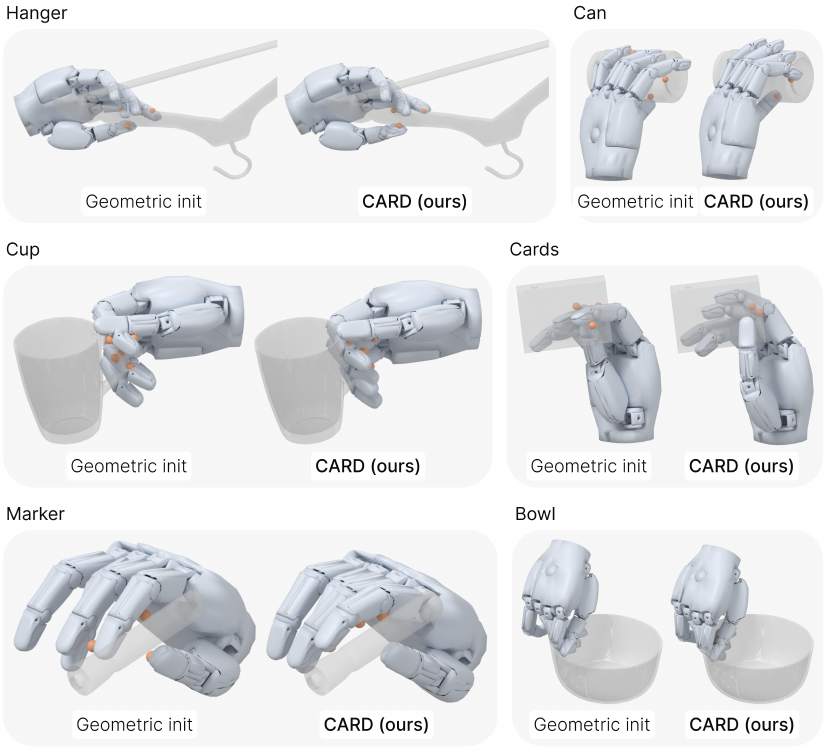}
    \caption{\textbf{Initializer vs.\ solution} on six video clips (hanger, can, cup, cards, marker, bowl): the geometric initializer $q^{\mathrm{init}}$ (left of each pair) seats the hand near the \fcas. The contact solve (right) closes the last millimeters. Orange spheres are the \fcas.}
    \label{fig:init}
\end{figure}

\section{Additional Vid-HACO Details and Analysis}
\label{app:vidhaco}
The model and baseline comparison are presented in \cref{sec:method:vidhaco,sec:exp:vidhaco,tab:contact}. Here we provide the training-data breakdown and additional evaluation details.

\subsection{Training Data and Implementation}

\nbf{Data composition} \Cref{tab:vidhaco_data} lists the 12 training datasets and their splits. Of the 1.55M training frames, 71\% come from HOT3D and ARCTIC's allocentric views, which HACO does not use. Labels are defined on the right-hand MANO mesh.

\begin{table}[t]
\centering\footnotesize
\caption{\textbf{Vid-HACO training data.} Contact: what the hand touches. View: egocentric, third-person (exo, including ARCTIC's eight allocentric cameras), or web images.}
\label{tab:vidhaco_data}
\setlength{\tabcolsep}{4pt}
\begin{tabular}{@{}lllrr@{}}
\toprule
Dataset & Contact & View & Train & Val. \\
\midrule
HOT3D~\citep{hot3d} & object & ego & 643{,}980 & 71{,}792 \\
ARCTIC~\citep{arctic}, allocentric & object & exo & 459{,}867 & 51{,}186 \\
ARCTIC~\citep{arctic}, egocentric & object & ego & 168{,}343 & 18{,}707 \\
HOI4D~\citep{hoi4d} & object & ego & 106{,}278 & 11{,}995 \\
DECAF~\citep{decaf} & face & exo & 58{,}829 & 6{,}493 \\
Hi4D~\citep{hi4d} & body & exo & 49{,}902 & 5{,}306 \\
InterHand2.6M~\citep{interhand} & hand & exo & 33{,}439 & 3{,}731 \\
H2O~\citep{h2ohands} & object & ego & 30{,}117 & 3{,}350 \\
EPIC-Contact~\citep{epiccontact} & object & ego & 2{,}035 & 233 \\
MOW~\citep{mow} & object & web & 307 & 43 \\
PROX~\citep{prox} & scene & exo & 157 & 21 \\
HIC~\citep{hic} & hand, object & exo & 99 & 16 \\
\midrule
Total & & & 1{,}553{,}353 & 172{,}873 \\
\bottomrule
\end{tabular}
\end{table}

\nbf{Implementation} The encoder, decoder, and two prediction heads in \cref{sec:method:vidhaco} have 411M parameters in total. The probability threshold of 0.52 is shared with HACO.

\subsection{Evaluation Details and False Contacts}
\label{app:contact}


\nbf{Benchmarks and metrics} The validation split (172.9k frames) measures in-distribution accuracy. HRDexDB~\citep{hrdexdb}, which no model saw in training, measures generalization: 434 egocentric videos of a hand grasping 94 objects (111.9k frames), with ground truth from the same 10\,mm rule. All models see the same hand crops and are thresholded at 0.52. Vertex precision, recall, and F1 are pooled over all vertices of all frames, so contact predicted on a contact-free frame lowers precision. AP is threshold-free, and frame F1 scores whether a frame is in contact at all.

\nbf{Fewer false contacts} Both HACO models over-predict contact, trading 11 points of precision for 3 to 5 points of recall on HRDexDB. Their class-balanced sampling removes nearly every contact-free frame from training, so they flag contact on every frame and score exactly the trivial always-in-contact frame F1, where Vid-HACO reaches 98.3 and 95.3. \textbf{This matters downstream: \ours turns contacting vertices into \fcas, so contact hallucinated during a reach pulls fingers onto empty space.}

\section{Geometric initialization}
\label{app:init}
\label{app:skel-init}

The initializer computes $q^{\mathrm{init}} = (t^{\mathrm{init}}, \omega^{\mathrm{init}}, \theta^{\mathrm{init}})$ from the human skeleton alone, in closed form, once per frame. No optimizer runs inside it. Let $g_0, \ldots, g_{20}$ be the MANO keypoints ($g_0$ the wrist, then MCP, PIP, DIP and TIP of each finger), let $\mathcal{B}$ be the set of hand bones, bone $k$ running from keypoint $i_k$ to $j_k$ with unit direction $\hat{e}^{\mathrm{H}}_k = (g_{j_k} - g_{i_k}) / \|g_{j_k} - g_{i_k}\|$, and let $\hat{e}_k(q)$ be the direction of the corresponding robot link, as in \cref{eq:style}. Let $\mathcal{P}$ be the pairs of adjacent bones $(k, l)$ along one finger, meeting at a joint, with the palm segment $g_0 \to g_{\mathrm{MCP}}$ as the first bone of every chain.

\nbf{Flexion} The robot's finger joints are hinges, so the angle between two adjacent robot links \emph{is} the hinge angle between them. Matching relative bone directions between adjacent bones therefore has a closed-form solution at every flexion hinge $j(k,l)$ of the index, middle, ring and pinky,
\begin{equation}
\label{eq:init-flex}
\theta^{\mathrm{init}}_{j(k,l)} = -\arccos\big(\hat{e}^{\mathrm{H}}_k \cdot \hat{e}^{\mathrm{H}}_l\big), \qquad (k,l) \in \mathcal{P},
\end{equation}
with the sign of the robot's flexion convention. The human bend is copied exactly. An attenuated copy compounds along the chain into a link-direction error (\cref{app:ablation}).

\nbf{Abduction} The component \cref{eq:init-flex} discards is the spread in the palm plane. With $\Pi$ the projection onto that plane, the abduction hinge of finger $i$ receives the signed angle between its proximal bone and the middle finger's, positive toward the thumb,
\begin{equation}
\label{eq:init-spread}
\theta^{\mathrm{init}}_{\mathrm{AA},i} = \angle\big(\Pi\,\hat{e}^{\mathrm{H}}_{\mathrm{prox},i},\ \Pi\,\hat{e}^{\mathrm{H}}_{\mathrm{prox},\mathrm{mid}}\big).
\end{equation}

\nbf{Wrist rotation} The palm frame $W = [\hat{f},\ \hat{a},\ \hat{f} \times \hat{a}]$ has $\hat{f}$ pointing from the wrist to the middle knuckle and $\hat{a}$ from the index to the pinky knuckle, made orthogonal to $\hat{f}$. $W^{\mathrm{H}}$ is built from the keypoints each frame and $W^{\mathrm{R}}$ once from the robot at rest, from its hand centre and the same three knuckles:
\begin{equation}
\label{eq:init-wrist}
R^{\mathrm{init}} = W^{\mathrm{H}}\,(W^{\mathrm{R}})^{\top}, \qquad \omega^{\mathrm{init}} = \log R^{\mathrm{init}}.
\end{equation}
Finger errors cannot rotate the hand.

\nbf{Thumb} The thumb admits neither \cref{eq:init-flex} nor \cref{eq:init-spread}: the human spreads it by about $50^\circ$ at the base where the robot's abduction hinge allows $20^\circ$, and the two chains' axes differ. Its five joints are instead an affine function of the human thumb geometry expressed in the palm frame,
\begin{equation}
\label{eq:init-thumb}
\begin{aligned}
\theta^{\mathrm{init}}_{\mathrm{thumb}} &= A\,x + b, \\
x &= (R^{\mathrm{init}})^{\top}\big[\hat{e}^{\mathrm{H}}_{\mathrm{CMC}\to\mathrm{MCP}},\ \hat{e}^{\mathrm{H}}_{\mathrm{MCP}\to\mathrm{IP}}, \\
&\qquad\qquad\quad\ \hat{e}^{\mathrm{H}}_{\mathrm{IP}\to\mathrm{TIP}},\ \hat{n}^{\mathrm{H}}_{\mathrm{thumb}}\big] \in \mathbb{R}^{12},
\end{aligned}
\end{equation}
with $\hat{n}^{\mathrm{H}}_{\mathrm{thumb}}$ the MANO thumb-pad normal of \cref{eq:style}. The constants $A \in \mathbb{R}^{5\times 12}$ and $b \in \mathbb{R}^5$ are fitted once, by ridge regression over 720 frames of 20 video clips, to the minimizer of the weighted least squares $\sum_k w_k^2 \|\hat{e}_k(\theta) - \hat{e}^{\mathrm{H}}_k\|^2 + w_{\mathrm{n}}^2 \|\hat{b}(\theta) - \hat{n}^{\mathrm{H}}_{\mathrm{thumb}}\|^2$ over the thumb joints within their limits (bone directions weighted $0.5$, $0.7$, $1.0$ from base to tip, pad normal $1.0$). On clips held out of the fit the map reproduces that minimizer to within half a degree. The per-frame least squares it replaces cost $6$\,ms, the map $0.03$\,ms.

\nbf{Wrist translation} With $a_i(\theta)$ the robot's pad apex of finger $i$ in the wrist frame, as in \cref{eq:style}, the translation seats the centroid of the copied fingertips on the centroid of the human fingertips,
\begin{equation}
\label{eq:init-seat}
t^{\mathrm{init}} = \frac{1}{5}\sum_{i=1}^{5} g_{\mathrm{TIP}_i} \;-\; R^{\mathrm{init}}\,\frac{1}{5}\sum_{i=1}^{5} a_i(\theta^{\mathrm{init}}),
\end{equation}
and $\theta^{\mathrm{init}}$ is clipped to the joint limits. The wrist keypoint is deliberately not used for position: the robot's fingers are about $40$\,mm longer than the human's, so matching fingertips rather than wrists places the robot palm $25$--$36$\,mm behind the human palm on the video clips, which is what lets the pads start near the \fcas. A rigid fit of rotation and translation to the five fingertips was tried and rejected, since five near-coplanar points determine a rotation poorly.

The initializer is deterministic, so a steady source gives a steady start, and it costs under a millisecond per frame. It enters the solve only as half of the sampler's warm start. None of the cost groups is evaluated inside it, and the style group scores the same quantities---pad apex, pad normal, link direction---in task space during the solve. \Cref{fig:init} shows the initializer beside the solved pose. We also tested the converse designs: a per-finger least squares on the style terms as the initializer places fingertips closer but contorts the chains, and the final solve is worse on every contact metric. The initializer's joint copy used as the free-finger reward in place of the style group cannot place fingertips and loses contact rate. The division of labour between a direction-exact initializer and a task-space style reward is therefore kept.

\section{Additional \ours implementation details}
\label{app:details}

\begin{figure*}[t]
    \centering
    \includegraphics[width=\linewidth]{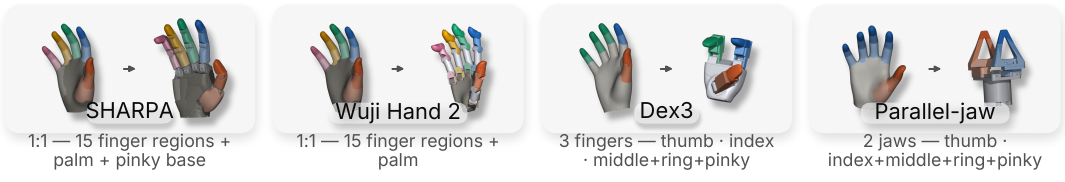}
    \caption{\textbf{Semantic maps.} The human hand (left of each pair) and the robot it is mapped to, colored by semantic region: the same color marks a human region and the robot link that receives its contacts. Skin marks human regions the embodiment has no link for, and grey marks unmapped robot bodies. SHARPA Wave and Wuji Hand~2 take the 15 finger regions and the palm one-to-one, and SHARPA also the pinky base. The Unitree Dex3-1 folds middle, ring, and pinky onto its third finger, and the parallel-jaw gripper assigns the thumb to one jaw and the four fingers to the other.}
    \label{fig:semantic_map}
\end{figure*}

\begin{figure}[t]
    \centering
    \includegraphics[width=0.7\linewidth]{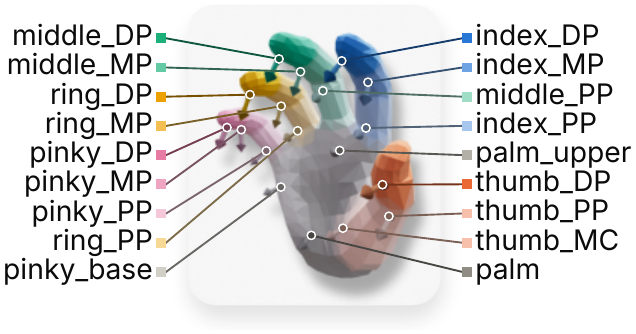}
    \caption{\textbf{Semantic regions.} The 18 regions shared by every source and robot, three per finger, the palm, the upper palm, and the pinky base, each with its region center of pressure.}
    \label{fig:regions}
\end{figure}

\nbf{Cross-embodiment semantic maps} A semantic map lists, for every human region of \cref{fig:regions}, the robot body that receives its \fcas and the face of that body that grasps. Nothing else changes between embodiments: the same anchors, cost groups, and weights (\cref{tab:solver}) drive every hand. \Cref{fig:semantic_map} shows the four maps we instantiate. SHARPA Wave and Wuji Hand~2 are five-fingered hands with three links per finger, so their maps are one-to-one: each of the 15 finger regions goes to the robot link of the same name (distal region to fingertip pad, middle to middle link, proximal to proximal link) and the palm to the palm body, and SHARPA additionally receives the pinky base on its metacarpal body. The Unitree Dex3-1 has three fingers, so its map is many-to-one: the human thumb maps to its thumb and the index to its index, while the middle, ring, and pinky regions all map to its third finger. Distal and middle regions land on the matching tip and middle pads, and when several of the grouped fingers touch at once their anchors are merged into one by a force-weighted centroid. The parallel-jaw gripper has two pads: the thumb maps to one jaw and the four fingers to the other, with the distal, middle, and proximal regions landing on the tip, middle, and base zones of the jaw, so the within-link location is still carried. Human regions without a counterpart, such as the palm on the Dex3-1 and on the gripper, carry no \fca.

\nbf{Cost groups in full} The style group of \cref{sec:method:retarget} reads
\begin{equation}
\label{eq:style}
\begin{aligned}
E_{\mathrm{style}}(q) ={}& w_{\mathrm{tip}}^2 \sum_{i=1}^{5} \big\| a_i(q) - a_i^{\mathrm{H}} \big\|^2 \\
&+ w_{\mathrm{n}}^2 \sum_{i=1}^{5} \big( 1 - \hat{b}_i(q) \cdot \hat{n}_i^{\mathrm{H}} \big) \\
&+ w_{\mathrm{b}}^2 \sum_{\ell} \big( 1 - \hat{e}_\ell(q) \cdot \hat{e}_\ell^{\mathrm{H}} \big),
\end{aligned}
\end{equation}
where $a_i(q)$ is the pad apex of finger $i$ and $a_i^{\mathrm{H}}$ the palmar apex of the matching MANO fingertip, $\hat{b}_i(q)$ the pad's belly axis and $\hat{n}_i^{\mathrm{H}}$ the MANO pad normal, and $\hat{e}_\ell$, $\hat{e}_\ell^{\mathrm{H}}$ the long axes of finger region $\ell$ on robot and human. The smoothness and collision groups are
\begin{equation}
\label{eq:smooth}
E_{\mathrm{smooth}}(q) = w_{\theta}^2 \big\| \theta - \theta^{-} \big\|^2 + w_{t}^2 \big\| t - t^{-} \big\|^2 + w_{\omega}^2 \big\| \omega \ominus \omega^{-} \big\|^2,
\end{equation}
\begin{equation}
\label{eq:collision}
\begin{aligned}
E_{\mathrm{collision}}(q) ={}& w_{\mathrm{self}}^2 \sum_{(i,j)} \max\!\big(0,\, m - h_{ij}(q)\big)^2 \\
&+ w_{\mathrm{obj}}^2 \sum_{s} \max\!\big(0,\, \delta_s(q) - \epsilon\big)^2 ,
\end{aligned}
\end{equation}
with $\ominus$ the rotation difference in the axis--angle chart nearest $\omega^{-}$, $h_{ij}(q)$ the capsule--capsule gap of finger pair $(i,j)$ with soft margin $m$, and $\delta_s(q)$ the penetration depth of hand-surface sample $s$ with allowance $\epsilon$. The two hard tests reject a candidate when $\max_s \delta_s(q) > \delta_{\max}$ or the palm--finger gap falls below $m$. Finger--finger pairs keep only the soft margin, since a valid tripod legitimately overlaps two fingertip capsules.

\begin{figure*}[t]
    \centering
    \includegraphics[width=\linewidth]{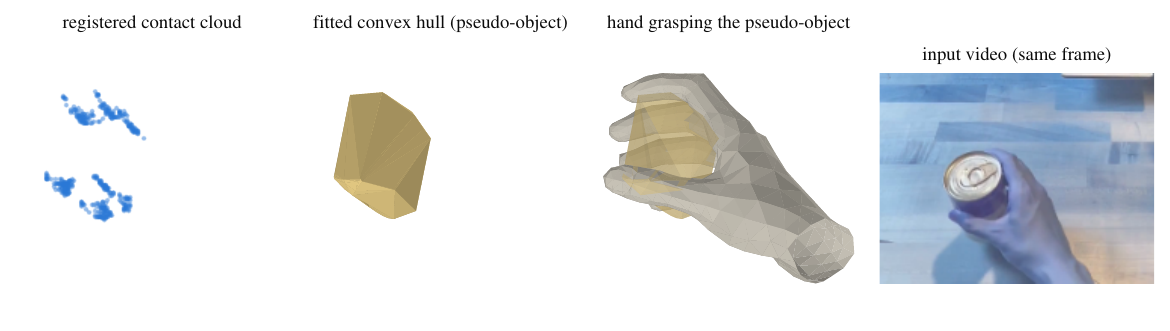}
    \caption{\textbf{Pseudo-object for video.} Left to right: the \fcas of every frame of a video clip, mapped into the wrist frame and accumulated (registered contact cloud); their convex hull, the pseudo-object that serves as the collision field; the MANO hand at the peak-contact frame grasping the translucent hull; and the input video at the same frame. The first three panels are drawn in the camera frame of that frame, so their orientation matches the video.}
    \label{fig:pseudoobject}
\end{figure*}

\nbf{Pseudo-object} The video and glove sources have no object mesh, so the object field of \cref{eq:collision} is built from the demonstration itself (\cref{fig:pseudoobject}). The positions $c_f$ of the active \fcas of every frame, palm excluded, are mapped into that frame's wrist frame and accumulated over the clip, and their convex hull is the \emph{pseudo-object} whose signed-distance field supplies $\delta_s(q)$. The hull covers only the contacted part of the object, a wedge of the can in \cref{fig:pseudoobject}, which is exactly the region the fingers must stay out of. Since the \fcas sit on the object surface, the demonstrating hand wraps the hull with little penetration. GRAB clips are solved the same way: the pseudo-object is the collision field and the GRAB mesh is used only for evaluation. Our own motion-capture clips, whose object is tracked live during capture, use the tracked mesh with the larger barrier weight of \cref{tab:solver}.

\nbf{Sampler} At iteration $i$ the sampler draws $N$ candidates $q_j = \bar{q}^{(i)} + \xi_j$, $\xi_j \sim \mathcal{N}(0, \mathrm{diag}(\sigma_i^2))$, keeps the elite set $\mathcal{E}_i$ of the lowest-cost $5\%$, and updates
\begin{equation}
\label{eq:mppi}
\begin{aligned}
\bar{q}^{(i+1)} &= \sum_{j \in \mathcal{E}_i} \frac{\exp(-\tilde{E}_j / \eta)}{\sum_{k \in \mathcal{E}_i} \exp(-\tilde{E}_k / \eta)}\, q_j, \\
\sigma_i &= \beta^{\,i} \sigma_0,\qquad \beta = \gamma^{1/K},
\end{aligned}
\end{equation}
where $\tilde{E}_j$ is the cost standardized over the elite set, so a uniform rescaling of all weights leaves the update unchanged, and the per-DoF noise anneals from $\sigma_0$ to $\gamma\sigma_0$ over $K$ iterations. The warm start is $\bar{q}^{(0)} = \tfrac{1}{2} q^{-} + \tfrac{1}{2} q^{\mathrm{init}}$. The incumbent is always re-evaluated and the best candidate seen is returned.

\nbf{Solver settings} \cref{tab:solver} lists the weights of \cref{eq:contact,eq:style,eq:smooth,eq:collision} and the sampler settings. One configuration serves the video, glove, and motion-capture sources.

\begin{table*}[t]
\centering\footnotesize
\caption{\textbf{Solver settings} shared by the video, glove, and motion-capture sources. Weights multiply the residual before squaring, as written in \cref{eq:contact,eq:style,eq:smooth,eq:collision}. Residual units differ across groups, so weights are not comparable by magnitude.}
\label{tab:solver}
\begin{tabular}{@{}llll@{}}
\toprule
Group & Term & Symbol & Value \\
\midrule
Contact & reach and cone, \cref{eq:contact} & $w_{\mathrm{c}}$ & 40 \\
 & direction price & $\lambda,\ r$ & 2,\ 10\,mm \\
 & correspondence blend & $\alpha_f$ & 0.15 (distal, middle), 0 (proximal) \\
 & contact stability & $w_{\mathrm{fix}}$ & 2.5 \\
 & force weighting & $\rho_f$ & $1 + \tfrac{1}{2} F_f / F_{\max} \in [1, 1.5]$ \\
\midrule
Style & fingertip apex & $w_{\mathrm{tip}}$ & 3 \\
 & pad normal & $w_{\mathrm{n}}$ & 60 \\
 & bone direction & $w_{\mathrm{b}}$ & 20 \\
\midrule
Smoothness & joints & $w_{\theta}$ & 0.1 \\
 & wrist translation & $w_{t}$ & $\sqrt{5^2 + 10^2}$\textsuperscript{$\dagger$} \\
 & wrist rotation & $w_{\omega}$ & 10 \\
\midrule
Collision & self-collision weight, margin & $w_{\mathrm{self}},\ m$ & 1.5,\ 6\,mm \\
 & object barrier weight, allowance & $w_{\mathrm{obj}},\ \epsilon$ & 3.5 (pseudo-object), 14 (motion capture, tracked mesh),\ 2\,mm \\
 & hard penetration floor & $\delta_{\max}$ & 3\,mm \\
\midrule
Sampler & candidates, iterations & $N,\ K$ & 1024,\ 16 \\
 & elite fraction, temperature & ---,\ $\eta$ & 5\%,\ 0.1 \\
 & initial noise (translation, rotation, joints) & $\sigma_0$ & 2\,mm,\ 0.016\,rad,\ 0.05\,rad \\
 & final noise scale & $\gamma$ & 0.2 \\
 & velocity caps (joints, translation) & --- & 0.25\,rad,\ 20\,mm per frame \\
 & warm start & $\bar{q}^{(0)}$ & $\tfrac{1}{2} q^{-} + \tfrac{1}{2} q^{\mathrm{init}}$ \\
\midrule
Post-processing & Savitzky--Golay window, order & --- & 15 frames,\ 3 \\
\bottomrule
\end{tabular}
\par\smallskip\raggedright\footnotesize\textsuperscript{$\dagger$}The implementation applies a weight of 5 to the translation step and a weight of 10 to the full 6-DoF base step. The latter also supplies $w_{\omega}$.
\end{table*}

\section{Ablations}
\label{app:ablation}

All ablations use 20 of the monocular-video recordings of \cref{tab:contact_quality} with the \fcas exactly as released by the upstream HACO pipeline, whose conservative gating leaves 4{,}297 anchor-frames. Rates are pooled over anchor-frames, Gap is the per-clip median and Pen.\ the per-clip maximum penetration, averaged over clips, and the full method is the mean of two sampler seeds (spread at most 1.1 points on rates, 0.1\,mm on distances).

\nbf{Initializer and barrier} \Cref{tab:ablation} removes the geometric initializer so the sampler starts from the previous frame alone, reverts to the earlier initializer (flexion copied at a gain of 0.7 with a per-frame least-squares thumb), and removes the soft penetration barrier. The initializer matters more than any single reward term. Without it the sampler still finds the \fcas on half the frames, but its worst misses grow to 3\,cm and penetration to 6\,mm: the warm start carries the fingertip information the sampler's small noise cannot recover. The earlier initializer lands 13 points below the current one. Its attenuated copy compounds along the chain to a $20^\circ$ link error, whereas the exact copy leaves $8^\circ$. The barrier costs 5.5 points of contact and keeps the deepest incursion at 4\,mm rather than 6.

\begin{table*}[t]
    \centering
    \caption{\textbf{Initializer and penetration barrier} on the 20-clip video bed. Rates in percent, Gap median and Pen.\ maximum in mm. Best in bold.}
    \label{tab:ablation}
    \footnotesize
    \setlength{\tabcolsep}{6pt}
    \begin{tabular}{lccccc}
    \toprule
    Variant & Contact $\uparrow$ & Region\,$|$\,contact $\uparrow$ & In-cone $\uparrow$ & Gap (mm) $\downarrow$ & Pen.\ (mm) $\downarrow$ \\
    \midrule
    \ours\ (full) & \textbf{64.7} & 97.4 & \textbf{82.2} & \textbf{2.0} & 4.2 \\
    $-$ geometric initializer & 52.3 & \textbf{98.8} & 79.0 & 2.5 & 6.0 \\
    previous initializer (gain 0.7) & 51.7 & 97.2 & 81.4 & 2.3 & \textbf{3.4} \\
    $-$ penetration barrier & 59.2 & 97.6 & 81.5 & \textbf{2.0} & 6.0 \\
    \bottomrule
    \end{tabular}
\end{table*}

\nbf{Contact terms} \Cref{tab:contact_ablation} switches the ingredients of the contact group \cref{eq:contact} on and off and removes the style group \cref{eq:style}. Every variant keeps the anti-slip term, the force weights, the smoothness and collision groups, the initializer, and the sampler budget of \cref{tab:solver}, and the contact point is always the nearest pad sample. \emph{Reach only} sets $\alpha_f{=}0$ and $\lambda{=}0$, a pure pull of the nearest sample onto the \fca. \emph{No correspondence} restores the cone but keeps $\alpha_f{=}0$. \emph{Correspondence only} is the opposite corner, $\alpha_f{=}1$ and $\lambda{=}0$. \emph{No cone} is the full method with $\lambda{=}0$, and \emph{no style} the full method with $w_{\mathrm{tip}}{=}w_{\mathrm{n}}{=}w_{\mathrm{b}}{=}0$.

Three things follow. The friction-cone term buys direction and pays for it in contact: every variant without it sits at 53--58\% in-cone, every variant with it at 80--82\%, but since the cone can only act by rotating the finger, the rotated pad reaches 4 points fewer \fcas (69.1 without the cone vs.\ 64.7 with it). We accept that trade, because a contact that cannot carry the demonstrated push is not the contact the human made. The correspondence term buys contact in either regime: adding $p^{\mathrm{H}}_f$ at $\alpha_f{=}0.15$ raises the contact rate from 58.4 to 64.7 with the cone and from 62.7 to 69.1 without it, while letting it own the position entirely ($\alpha_f{=}1$) over-constrains the pad (50.4 and 59.1), so the blend must stay small. The style group costs 5 points on the anchored links (59.9 vs.\ 64.7) but is what shapes the free fingers.

\begin{table*}[t]
\centering\small
\caption{\textbf{Contact-term ablation} on the 20-clip video bed. $\alpha_f$ is the correspondence blend and $\lambda$ the friction-cone weight of \cref{eq:contact}, and $E_{\mathrm{style}}$ is the style group \cref{eq:style}. Gap median and Pen.\ maximum in mm. The contact point is the nearest pad sample throughout. Best in bold.}
\label{tab:contact_ablation}
\footnotesize
\setlength{\tabcolsep}{6pt}
\begin{tabular}{@{}llccccc@{}}
\toprule
Variant & Setting & Contact $\uparrow$ & Region\,$|$\,contact $\uparrow$ & In-cone $\uparrow$ & Gap $\downarrow$ & Pen.\ $\downarrow$ \\
\midrule
\ours\ (full) & $\alpha_f{=}0.15,\ \lambda{=}2$ & 64.7 & 97.4 & \textbf{82.2} & 2.0 & 4.2 \\
\midrule
Reach only & $\alpha_f{=}0,\ \lambda{=}0$ & 62.7 & 98.1 & 57.5 & 1.7 & 4.4 \\
No correspondence & $\alpha_f{=}0,\ \lambda{=}2$ & 58.4 & \textbf{98.5} & 81.3 & 2.1 & 4.5 \\
Correspondence only & $\alpha_f{=}1,\ \lambda{=}0$ & 59.1 & 97.6 & 53.3 & 1.9 & 4.2 \\
Correspondence $+$ cone & $\alpha_f{=}1,\ \lambda{=}2$ & 50.4 & 97.8 & 80.3 & 2.5 & \textbf{4.0} \\
No cone & $\alpha_f{=}0.15,\ \lambda{=}0$ & \textbf{69.1} & 97.3 & 54.6 & \textbf{1.5} & 4.2 \\
No style & $E_{\mathrm{style}} \equiv 0$ & 59.9 & 96.9 & 80.1 & 2.1 & 4.1 \\
\bottomrule
\end{tabular}
\end{table*}

\section{Tactile tracking controller}
\label{app:tracker}

\nbf{Formulation} \Cref{fig:tracker_diagram} and \cref{eq:controller} in the main text define the loop, one per finger $f$ at $100$\,Hz, with the kinematic path in blue and the force path in orange; the contact state machine below is not drawn. The SHARPA Wave measures the contact wrench at each fingertip, from which we take the pad's surface-normal force $F^{\mathrm{meas}}_f$, clamped at zero so a dropped frame reads as free space, and the force error $e_f = F^{\mathrm{des}}_f - F^{\mathrm{meas}}_f$ is low-passed at $10$\,Hz to $\bar{e}_f$. The admittance is a rate-limited integrator with a small proportional term,
\begin{equation}
\label{eq:admittance}
\dot{\kappa}^{\mathrm{I}}_f = \mathrm{clip}\big(k_i\,\bar{e}_f,\, \pm\dot{\kappa}_{\max}\big), \qquad
\kappa_f = \mathrm{clip}\big(\kappa^{\mathrm{I}}_f + k_p\,\bar{e}_f,\, 0,\, 1\big),
\end{equation}
with $k_i = 1.0$ curl\,/\,(N$\cdot$s), $k_p = 0.05$ curl\,/\,N, $\dot{\kappa}_{\max} = 6\,\mathrm{s}^{-1}$, and $k_{\mathrm{ff}} = 0.3$ in \cref{eq:controller}. The rate limit and the clamp supply both the anti-slam approach limit and anti-windup, and the proportional term sits outside the integrator state so it cannot wind up. The curl deflects only the finger's flexion joints, along the fixed unit synergy $\hat{\psi}_f$, toward closure; the Jacobian in $\tau_{\mathrm{ff}}$ is evaluated at $q^{\mathrm{cmd}}$. Five independent loops run on the $22$-DoF hand, and the gains are dimensionless in curl, so they carry between simulation and hardware unchanged.

\nbf{Contact state machine} Following the contact-event grasp controllers of Romano et al.~\citep{romano2011}, the contact state machine enters on $0.05$\,N confirmed for one frame and exits below $0.01$\,N sustained for $0.6$\,s. During that wait the integrator freezes rather than continuing to integrate the full set-point, which is what prevents a finger from crushing shut on a momentary sensor dropout. On declared loss the applied curl relaxes toward the retargeted trajectory with a $1.0$\,s time constant, and a re-touch resumes from the decayed curl rather than from the pre-loss integrator. The feedforward torque uses $k_{\mathrm{ff}} = 0.3$, is armed only on confirmed contact, and is scaled uniformly---not clipped per element, which would distort the torque distribution across a finger's joints---to $\lVert\tau_{\mathrm{ff}}\rVert_\infty \le 0.15$\,N$\cdot$m. The curl deflection is capped at $0.10$\,rad per joint beyond the retargeted target and desired forces at $10$\,N per finger.

\nbf{Friction identification} The feedforward torque only reaches the fingertip if joint friction is known, so we identified it before the experiments. Joint friction is compensated as a per-joint, per-direction Coulomb-plus-viscous model with a stiction breakaway threshold, fit from constant-velocity sweeps and breakaway ramps in simulation on the hand's own model and validated against its ground-truth friction parameters. It is evaluated on commanded rather than measured velocity.

\nbf{Arm} The arm tracks the retargeted wrist by differential inverse kinematics, smoothed with a zero-phase Savitzky--Golay filter (7-frame window, order 2) under a $90^\circ$/s per-joint velocity ceiling, then interpolated to a $500$\,Hz stream.

\begin{figure*}[t]
    \centering
    \includegraphics[width=0.8\linewidth]{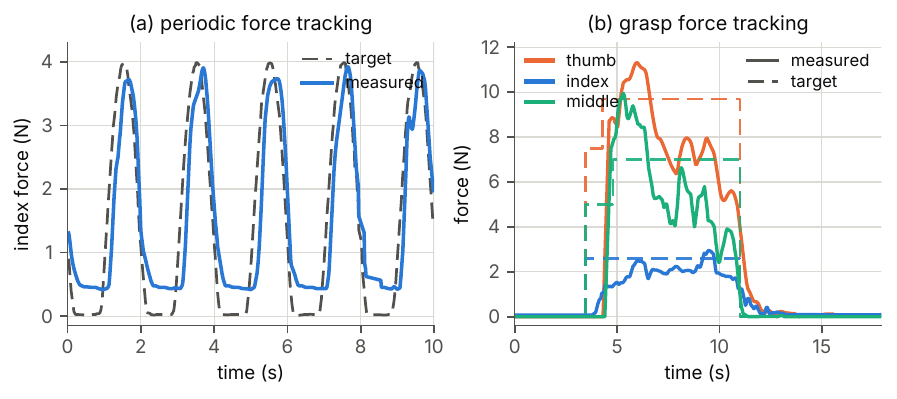}
    \caption{\textbf{Force tracking on the SHARPA Wave.} (a) The index finger follows a $0$--$4$\,N periodic contact force target against a fixed object: the measured force reaches the peaks within a few tenths of a newton and lags the target by about $0.2$\,s. The floor near $0.5$\,N is the preload the state machine keeps while in contact. (b) A three-finger grasp with step targets during mustard bottle pick-up replay (thumb $9.7$\,N, middle $7$\,N, index $2.6$\,N): the forces rise to their set-points within a second of the step, the thumb overshoots while the object settles, and release returns every finger to zero within about a second.}
    \label{fig:force_tracking}
\end{figure*}

\nbf{Results} \Cref{fig:force_tracking} shows the closed-loop force tracking on a fixed object and in a three-finger grasp. The task-level effect of the loop is the replay study in \cref{sec:exp:real,fig:tactile_ablation}, which compares identical retargeted trajectories with and without the tracker.

\section{Metric definitions}
\label{app:metrics}

\nbf{Evaluation metrics} We keep the notation of \cref{sec:method:uhcr}: $\mathcal{A}$ is the set of \emph{anchor-frames} --- one \fca $C_f = (c_f, \hat{d}_f, F_f)$ of one frame --- and $N_{\mathrm{a}} = |\mathcal{A}|$ their number. For an anchor-frame $a \in \mathcal{A}$ with assigned link $f$, $M_f$ is the link's visual geometry in its body frame, $\mathcal{S}^{\mathrm{pal}}_f \subset \mathcal{S}_f$ the pad samples on its palmar face, and $s^{*}_f$ the sample nearest the \fca. Every method is graded against the identical stored \fcas, so only the hand pose varies between columns. We use the following five metrics:
\begin{itemize}
    \item \textbf{Contact rate:} A contact is considered \emph{established} when the \fca lies inside the assigned link, or lies within $\delta_{\mathrm{c}} = 2$\,mm of its surface and is reached with the palmar face of the pad. CR is the fraction of anchor-frames whose contact is established:
    \begin{equation}
    \begin{aligned}
    \chi_a &\;=\; \mathbf{1}\!\left[c_f\in M_f\right] \;\vee\; \mathbf{1}\!\left[\operatorname{dist}\!\left(c_f,\partial M_f\right)\le\delta_{\mathrm{c}}\right] \\
    &\qquad\cdot\mathbf{1}\!\left[s^{*}_f \in \mathcal{S}^{\mathrm{pal}}_f\right],\\
    \mathrm{CR} &\;=\; \frac{1}{N_{\mathrm{a}}}\sum_{a\in\mathcal{A}}\chi_a.
    \end{aligned}
    \label{eq:established}
    \end{equation}
    \item \textbf{Correct region:} A contact is only considered in the correct region when the contact is established, and the palmar pad nearest the anchor over the whole hand belongs to the demonstrated finger. RR is the fraction of anchor-frames in the correct region.
    \item \textbf{In-cone:} A contact is only considered in-cone when the contact is established, and the contact normal $\hat{n}(s^{*}_f(q))$ lies within the $45^\circ$ friction cone centered on the demonstrated push direction $\hat{d}_f$. The dot product is signed, so a normal facing away from the push fails. IC is the fraction of in-cone anchor-frames:
    \begin{equation}
    \mathrm{IC}=\frac{1}{N_{\mathrm{a}}}\sum_{a\in\mathcal{A}}\chi_a\,\mathbf{1}\!\left[\hat{n}(s^{*}_f(q))\cdot\hat{d}_f\ge\cos 45^\circ\right].
    \end{equation}
    \item \textbf{Contact gap:} The gap of an anchor-frame is the distance from the \fca to the assigned link's surface, in mm:
    \begin{equation}
    \mathrm{Gap}_a=\operatorname{dist}\!\left(c_f,\partial M_f\right).
    \end{equation}
    \item \textbf{Penetration:} The penetration of a frame is the deepest incursion of the hand surface into the object, in mm, with $\delta_s$ the signed incursion of hand-surface sample $s$ into the object's signed-distance field:
    \begin{equation}
    \mathrm{Pen}=\max_s\,\delta_s(q).
    \end{equation}
    It is reported only where a ground-truth tracked object mesh exists (GRAB, motion capture); the video and glove clips carry no object ground truth.
\end{itemize}

\section{Contact fidelity by demonstration source}
\label{app:by_source}

\Cref{tab:by_source} splits the clips of \cref{tab:contact_quality} by how the contact signal was obtained: measured taxel pressure on the glove, contact inferred by Vid-HACO on video, proximity-thresholded contact on our motion capture, and GRAB, whose contact comes from the ground-truth meshes. \ours follows the quality of the contact signal: it is strongest where contact is measured (89.1\% contact, 52.2\% in-cone, 1.5\,mm gap on the glove), then on tracked motion capture (76.1\%), weaker on video (61.4\%), where Vid-HACO infers the contact and a wrong label places the robot pad on empty space, and weakest on GRAB (43.1\%), whose proximity labels carry no pressure and no true center of pressure. The kinematic baselines stay at 1--10\% on every source and move the other way, best on video (7.6--9.7\%) and worst on the glove (1.3--1.8\%), which is what methods that track keypoints should do: they gain nothing from a better contact signal. The geometric initializer, \ours's own deterministic warm start, already exceeds every baseline on every source (12.2--43.2\%): a human-isomorphic pose places centers of pressure inside the link far more reliably than fitting scaled fingertip targets does.

\begin{table*}[t]
    \centering
    \small
    \setlength{\tabcolsep}{6pt}
    \caption{\textbf{Contact fidelity by source.} The clips of \cref{tab:contact_quality} grouped by contact signal. Rates in percent, pooled over anchor-frames; millimeter rows are the median over clips of the per-clip median. Penetration is reported only where a tracked object mesh exists. \textbf{Bold}: best per source and metric.}
    \label{tab:by_source}
    \begin{tabular}{lccccc}
        \toprule
        & \ours & DexPilot & AnyTeleop-K & AnyTeleop-P & \ours \\
        & & \citep{dexpilot} & \citep{anyteleop} & \citep{anyteleop} & (init only) \\
        \midrule
        \multicolumn{6}{l}{\emph{Tactile glove, measured pressure (18 clips, 5{,}464 anchor-frames)}} \\
        \quad Contact rate (\%) $\uparrow$ & \textbf{89.1} & 1.8 & 1.3 & 1.3 & 43.2 \\
        \quad Correct region (\%) $\uparrow$ & \textbf{87.8} & 1.8 & 1.3 & 1.3 & 43.2 \\
        \quad In-cone (\%) $\uparrow$ & \textbf{52.2} & 0.5 & 0.4 & 0.4 & 6.8 \\
        \quad Gap (mm) $\downarrow$ & \textbf{1.54} & 30.59 & 30.25 & 29.16 & 8.54 \\
        \midrule
        \multicolumn{6}{l}{\emph{Monocular video, inferred contact (20 clips, 5{,}259 anchor-frames)}} \\
        \quad Contact rate (\%) $\uparrow$ & \textbf{61.4} & 9.7 & 8.3 & 7.6 & 20.5 \\
        \quad Correct region (\%) $\uparrow$ & \textbf{60.1} & 9.5 & 8.0 & 7.5 & 19.2 \\
        \quad In-cone (\%) $\uparrow$ & \textbf{35.2} & 4.2 & 4.7 & 5.0 & -- \\
        \quad Gap (mm) $\downarrow$ & \textbf{2.40} & 9.99 & 10.80 & 13.19 & 5.76 \\
        \midrule
        \multicolumn{6}{l}{\emph{Motion capture, proximity contact (48 clips, 40{,}254 anchor-frames)}} \\
        \quad Contact rate (\%) $\uparrow$ & \textbf{76.1} & 4.0 & 3.6 & 3.8 & 12.2 \\
        \quad Correct region (\%) $\uparrow$ & \textbf{75.5} & 4.0 & 3.6 & 3.5 & 12.1 \\
        \quad In-cone (\%) $\uparrow$ & \textbf{45.1} & 1.2 & 1.1 & 2.6 & -- \\
        \quad Gap (mm) $\downarrow$ & \textbf{1.36} & 15.35 & 14.83 & 22.41 & 9.43 \\
        \midrule
        \multicolumn{6}{l}{\emph{GRAB, mesh-derived contact (100 clips, 99{,}969 anchor-frames)}} \\
        \quad Contact rate (\%) $\uparrow$ & \textbf{43.1} & 4.9 & 4.9 & 4.6 & 19.6 \\
        \quad Correct region (\%) $\uparrow$ & \textbf{40.6} & 4.8 & 4.9 & 4.5 & 17.8 \\
        \quad In-cone (\%) $\uparrow$ & \textbf{16.2} & 1.7 & 1.7 & 1.7 & 3.8 \\
        \quad Gap (mm) $\downarrow$ & \textbf{3.56} & 20.90 & 20.45 & 20.66 & 13.78 \\
        \quad Penetration (mm) $\downarrow$ & \textbf{7.96} & 12.71 & 12.79 & 13.02 & 11.15 \\
        \bottomrule
    \end{tabular}
\end{table*}

\section{\ours--physics}
\label{app:physics}

\nbf{Setup} SPIDER~\citep{spider} retargets in two stages. A kinematic stage fits the robot fingertips to the MANO keypoints by inverse kinematics and synthesizes contact targets from a replay of those keypoints; a physics stage rolls the hand and the object forward in MuJoCo with a sampling-based model-predictive controller that tracks the kinematic reference and rewards contact. \ours-physics keeps the physics stage and its budget (256 samples and 8 iterations per 2\,s segment, early-abort gate off so that every rollout covers the whole demonstration) and changes three things. (1)~The sampler is initialized with, and tracks, the \ours trajectory of \cref{sec:method:retarget} in place of the fingertip-IK reference; the object track and the resampling, padding, and smoothing of SPIDER's kinematic stage are kept, so both arms follow the identical object reference. (2)~SPIDER's fingertip contact term is replaced by our contact objective $E_{\mathrm{contact}}(q)$ of \cref{eq:contact}: the squared distance of each contacting pad to its contact force target enters the reward with weight 10 and the $45^\circ$ friction-cone term on the demonstrated push with weight 2, relative to SPIDER's tracking terms; the anti-slip term is omitted. (3)~The \fca come from our pipeline (\cref{app:data_unification}) with the geometric contact rule used for GRAB, a contact probability of $1 - d/9\,\mathrm{mm}$ at hand--object distance $d$, thresholded at $0.5$. SPIDER needs the object mesh and pose track to run at all, so the anchors add no information the baseline lacks.

\nbf{Data} 48 motion-capture demonstrations from our TableMocap sessions, including glove-instrumented grasps of a mustard bottle, each with a tracked object mesh (two seeds per arm), and 50 GRAB clips cut to the stored contact window plus 0.5\,s on either side (one seed).

\nbf{Scoring} SPIDER's own criterion, a final object pose within 10\,cm and $90^\circ$ of the reference, is met by rollouts that never touch an object that returns near its start, so we score each run post hoc from its trajectory. A run succeeds when it covers the whole demonstration, the object travels at least half the demonstrated distance and, if the demonstration lifts it by 5\,cm or more, rises by at least 3\,cm and half the demonstrated lift, and the peak object position error stays below 15\,cm. A task is solved when at least one seed succeeds; runs with invalid physics do not count. \Cref{tab:physics} reports task success. Per run, \ours-physics succeeds on 51\% of the motion-capture runs against 39\% for SPIDER and on 43\% of the GRAB runs against 10\%, with median peak object errors of 106 vs.\ 138\,mm and 120 vs.\ 274\,mm. The rollouts show the mechanism: the IK reference forms thumb--finger opposition only after the object has started to move, so the sampler must find the grasp on its own, whereas the \ours reference encloses the object at the demonstrated contact onset and the sampler only has to squeeze.